\documentclass[nohyperref]{article}

\usepackage{amsmath,amsfonts,bm}

\def\eqref#1{equation~\ref{#1}}
\def\1{\bm{1}}

\def\vh{{\bm{h}}}

\def\vs{{\bm{s}}}

\def\vx{{\bm{x}}}
\def\vy{{\bm{y}}}

\def\mA{{\bm{A}}}

\def\mC{{\bm{C}}}
\def\mD{{\bm{D}}}

\def\mH{{\bm{H}}}
\def\mI{{\bm{I}}}

\def\mL{{\bm{L}}}

\def\mP{{\bm{P}}}

\def\mS{{\bm{S}}}

\def\mW{{\bm{W}}}
\def\mX{{\bm{X}}}
\def\mY{{\bm{Y}}}

\DeclareMathAlphabet{\mathsfit}{\encodingdefault}{\sfdefault}{m}{sl}
\SetMathAlphabet{\mathsfit}{bold}{\encodingdefault}{\sfdefault}{bx}{n}

\def\gD{{\mathcal{D}}}
\def\gE{{\mathcal{E}}}

\def\gG{{\mathcal{G}}}

\def\gL{{\mathcal{L}}}

\def\gN{{\mathcal{N}}}

\def\gV{{\mathcal{V}}}

\def\sR{{\mathbb{R}}}

\usepackage{marginnote}
\usepackage{enumitem}
\usepackage[utf8]{inputenc} 
\usepackage[T1]{fontenc}  
\usepackage{url}       
\usepackage{booktabs}  
\usepackage{amsfonts} 
\usepackage{nicefrac}   
\usepackage{microtype} 
\usepackage{url,hyphenat}
\usepackage{caption,subfigure,graphicx,epstopdf,multirow,multicol,booktabs,verbatim,wrapfig,bm}
\usepackage{titlesec}
\usepackage{makecell}
\usepackage{xspace}
\usepackage{amsthm,amssymb,amsfonts,amsmath}
\usepackage{threeparttable}
\usepackage{listings}
\usepackage{pythonhighlight}
\usepackage{array}
\usepackage{bbm}
\usepackage{dsfont}
\usepackage{xcolor,colortbl}

\usepackage{booktabs}

\usepackage[colorlinks,
            linkcolor=red,
            anchorcolor=blue,
            citecolor=purple,
            ]{hyperref}

\newcommand{\chameleon}{\texttt{Chameleon}\xspace}
\newcommand{\squirrel}{\texttt{Squirrel}\xspace}
\newcommand{\actor}{\texttt{Actor}\xspace}
\newcommand{\cora}{\texttt{Cora}\xspace}
\newcommand{\citeseer}{\texttt{Citeseer}\xspace}
\newcommand{\pubmed}{\texttt{Pubmed}\xspace}
\newcommand{\computer}{\texttt{Amazon-Computer}\xspace}
\newcommand{\photo}{\texttt{Amazon-Photo}\xspace}
\newcommand{\cs}{\texttt{Coauthor-CS}\xspace}
\newcommand{\physics}{\texttt{Coauthor-Physics}\xspace}

\newcommand{\arxiv}{\texttt{Ogbn-Arxiv}\xspace}
\newcommand{\products}{\texttt{Ogbn-Products}\xspace}
\definecolor{backcolor}{RGB}{232, 242, 255}

\usepackage{microtype}
\usepackage{graphicx}
\usepackage{subfigure}
\usepackage{booktabs} % for professional tables

\usepackage{hyperref}

\usepackage[accepted]{icml2023}

\usepackage{amsmath}
\usepackage{amssymb}
\usepackage{mathtools}
\usepackage{amsthm}

\usepackage[capitalize,noabbrev]{cleveref}

\theoremstyle{plain}
\newtheorem{theorem}{Theorem}[section]

\newtheorem{lemma}[theorem]{Lemma}

\theoremstyle{definition}
\newtheorem{definition}[theorem]{Definition}

\theoremstyle{remark}
\newtheorem{remark}[theorem]{Remark}

\usepackage[textsize=tiny]{todonotes}

\icmltitlerunning{OrthoReg: Improving Graph-regularized MLPs via Orthogonality Regularization}
\newcommand{\modelname}{{\sc Ortho-Reg}\xspace}

\begin{document}

\twocolumn[
\icmltitle{OrthoReg: Improving Graph-regularized MLPs via \\ Orthogonality Regularization}



\icmlsetsymbol{equal}{*}

\begin{icmlauthorlist}
\icmlauthor{Hengrui Zhang}{uic}
\icmlauthor{Shen Wang}{amazon}
\icmlauthor{Vassilis N. Ioannidis}{amazon}
\icmlauthor{Soji Adeshina}{amazon}
\icmlauthor{Jiani Zhang}{amazon}
\icmlauthor{Xiao Qin}{amazon}
\icmlauthor{Christos Faloutsos}{cmu}
\icmlauthor{Da Zheng}{amazon}
\icmlauthor{George Karypis}{amazon}
\icmlauthor{Philip S. Yu}{uic}

\end{icmlauthorlist}

\icmlaffiliation{uic}{University of Illinois Chicago}
\icmlaffiliation{amazon}{Amazon Web Services}
\icmlaffiliation{cmu}{Carnegie Mellon University}
\icmlcorrespondingauthor{Philip S. Yu}{psyu@uic.edu}

\icmlkeywords{Machine Learning, ICML}

\vskip 0.3in
]



\printAffiliationsAndNotice{\icmlEqualContribution} % otherwise use the standard text.

\begin{abstract}
    Graph Neural Networks (GNNs) are currently dominating in modeling graph-structured data, while their high reliance on graph structure for inference significantly impedes them from widespread applications. By contrast, Graph-regularized MLPs (GR-MLPs) implicitly inject the graph structure information into model weights, while their performance can hardly match that of GNNs in most tasks. This motivates us to study the causes of the limited performance of GR-MLPs. In this paper, we first demonstrate that node embeddings learned from conventional GR-MLPs suffer from dimensional collapse, a phenomenon in which the largest a few eigenvalues dominate the embedding space, through both empirical observations and theoretical analysis. As a result, the expressive power of the learned node representations is constrained. We further propose \modelname, a novel GR-MLP model to mitigate the dimensional collapse issue. Through a soft regularization loss on the correlation matrix of node embeddings, \modelname explicitly encourages orthogonal node representations and thus can naturally avoid dimensionally collapsed representations. Experiments on traditional transductive semi-supervised classification tasks and inductive node classification for cold-start scenarios demonstrate its effectiveness and superiority.
\end{abstract}

\section{Introduction}\label{sec:intro}
\begin{figure}[h]
    \centering
    \includegraphics[width = 0.95\linewidth]{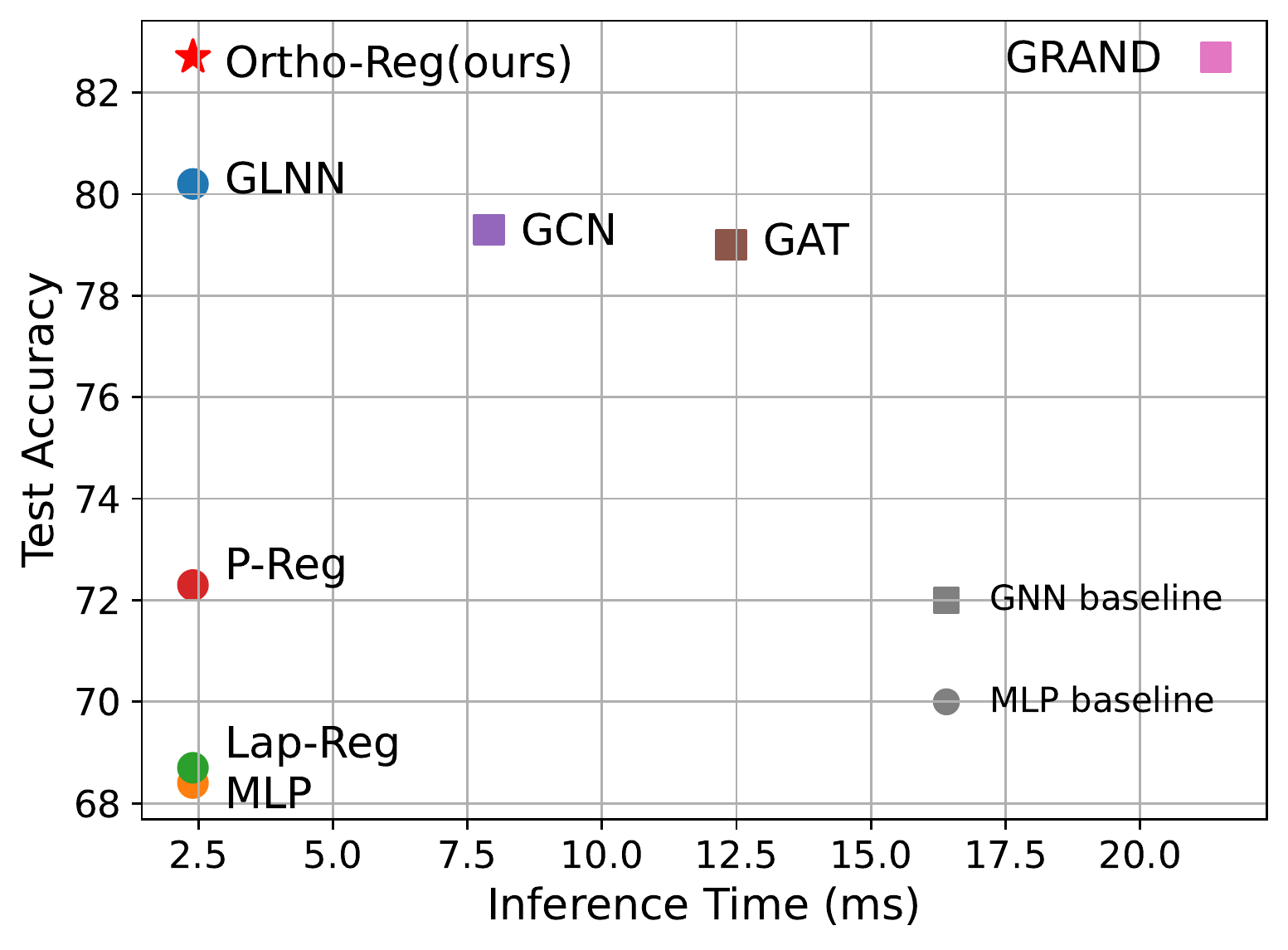}
    \caption{As an MLP model, our method performs even better than GNN models on \pubmed, but with a much faster inference speed. GRAND~\citep{grand} is one of the SOTA GNN models on this task. Circled markers denote MLP baselines, and squared markers indicate GNN baselines.}
    \label{fig:intro}
\end{figure}

Graph Machine Learning (GML) has been attracting increasing attention due to its wide applications in many real-world scenarios, like social network analysis~\citep{social_rec}, recommender systems~\citep{gcmc, danser}, chemical molecules~\citep{molclr, 3dinfomax} and biology structures. Graph Neural Networks (GNNs)~\citep{gcn, graphsage, gat, gin} are currently the dominant models for GML thanks to their powerful representation capability through iteratively aggregating information from neighbors. Despite their successes, such an explicit utilization of graph structure information hinders GNNs from being widely applied in industry-level tasks. On the one hand, GNNs rely on layer-wise message passing to aggregate features from the neighborhood, which is computationally inefficient during inference, especially when the model becomes deep ~\citep{glnn}. On the other hand, recent studies have shown that GNN models can not perform satisfactorily in cold-start scenarios where the connections of new incoming nodes are few or unknown~\citep{coldbrew}. By contrast, Multi-Layer Perceptrons (MLPs) involve no dependence between pairs of nodes, indicating that they can infer much faster than GNNs ~\citep{glnn}. Besides, they can predict for all nodes fairly regardless of the number of connections, thus can infer more reasonably when neighborhoods are missing ~\citep{coldbrew}. However, it remains challenging to inject the knowledge of graph structure information into learning MLPs.

One classical and popular method to mitigate this issue is Graph-Regularized MLPs (GR-MLPs in short). Generally, besides the basic supervised loss (e.g., cross-entropy),  GR-MLPs employ an additional regularization term on the final node embeddings or predictions based on the graph structure~\citep{lap-reg, label-propagation, p-reg, graphmlp}. Though having different formulations, the basic idea is to make node embeddings/predictions smoothed over the graph structure. Even though these GR-MLP models can implicitly encode the graph structure information into model parameters, there is still a considerable gap between their performance compared with GNNs~\citep{lap-reg, p-reg}. Recently, another line of work, GNN-to-MLP knowledge distillation methods (termed by KD-MLPs)~\citep{glnn, coldbrew}, have been explored to incorporate graph structure with MLPs. In KD-MLPs, a student MLP model is trained using supervised loss and a knowledge-distillation loss from a well-trained teacher GNN model. Empirical results demonstrate that with merely node features as input, the performance of KD-MLPs can still match that of GNNs as long as they are appropriately learned. However, the 2-step training of KD-MLPs is undesirable, and they still require a well-trained GNN model as a teacher. This motivates us to rethink the failure of previous GR-MLPs to solve graph-related applications and study the reasons that limit their performance.

\textbf{Presented work:} In this paper, we first demonstrate that node embeddings learned from existing GR-MLPs suffer from dimensional collapse~\citep{feature-decorrelation,understand-dimensional-collapse}, a phenomenon that the embedding space of nodes is dominated by the largest (a few) eigenvalue(s). Our theoretical analysis demonstrates that the dimensional collapse in GR-MLP is due to the irregular feature interaction caused by the graph Laplacian matrix (see Lemma~\ref{lemma:weight_matrix-shrink}). We then propose Orthogonality Regularization (\modelname in short), a novel GR-MLP model, to mitigate the dimensional collapse issue in semi-supervised node representation learning tasks. The key design of \modelname is to enforce an additional regularization term on the output node embeddings, making them \textbf{orthogonal} so that different embedding dimensions can learn to express various aspects of information. Besides, \modelname extends the traditional first-order proximity preserving target to a more flexible one, improving the model's expressive power and generalization ability to non-homophily graphs. We provide a thorough evaluation for \modelname on various node classification tasks. The empirical results demonstrate that \modelname can achieve competitive or even better performance than GNNs. Besides, using merely node features to make predictions, \modelname can infer much faster on large-scale graphs and make predictions more reasonable for new nodes without connections. In Fig.~\ref{fig:intro} we present the performance of \modelname compared with GNNs and other MLPs on \pubmed, where \modelname achieves SOTA performance with the fastest inference speed.

\textbf{We summarize our contributions as follows}:
\begin{itemize}
% [topsep=0in,leftmargin=em,wide=0em]
    \item[\textbf{1)}]
    We are the first to examine the limited representation power of existing GR-MLP models from the perspective of dimensional collapse. We provide theoretical analysis and empirical studies to justify our claims.
    \item[\textbf{2)}] To mitigate the dimensional collapse problem, we design a novel GR-MLP model named \modelname. \modelname encourages the node embeddings to be orthogonal through explicit soft regularization, thus can naturally avoid dimensional collapse. 
    \item[\textbf{3)}]  We conduct experiments on traditional transductive semi-supervised node classification tasks and inductive node classification under cold-start scenarios on public datasets of various scales. The numerical results and analysis demonstrate that by learning orthogonal node representations, \modelname can outperform GNN models on these tasks.
\end{itemize}
\vspace{-6pt}
\section{Backgrounds and Related Works}
\vspace{-6pt}
\subsection{Problem Formulation}
We mainly study a general semi-supervised node classification task on a single homogeneous graph where we only have one type of node and edge. We denote a graph by $\gG = (\gV, \gE)$, where $\gV$ is the node set, and $\gE$ is the edge set. For a graph with $N$ nodes (i.e., $|\gV| = N$), we denote the node feature matrix by $\mX \in \sR^{N \times D}$, the adjacency matrix by $\mA \in \sR^{N \times N}$. In semi-supervised node classification tasks, only a small portion of nodes are labeled, and the task is to infer the labels of unlabeled nodes using the node features and the graph structure. Denote the labeled node set by $\gV^{L}$ and the unlabeled node set by $\gV^{U}$, then we have $\gV^{L} \cap \gV^{U} = \varnothing$ and $\gV^{L} \cup \gV^{U} = \gV$. 

Denote the one-hot ground-truth labels of nodes by $\hat{\mY} \in \mathbb{R}^{N \times C}$, and the predicted labels by ${\mY}$. One can learn node embeddings $\mH$ using node features $\mX$ and adjacency matrix $\mA$, and use the embeddings to generate predicted labels $\hat{\mY}$. For example, GNNs generate node representations through iteratively aggregating and transforming the embeddings from the neighbors and could be generally formulated as $\mH = f_{\theta}(\mX, \mA)$. Then a linear layer is employed on top of node embeddings to predict the labels $ \mY = g_{\theta}(\mH)$. The model could be trained in an end-to-end manner by optimizing the cross-entropy loss between predicted labels and ground-truth labels of labeled nodes: $\mathcal{L}_{sup} = \bm{\ell}_{xent}(\mY^{L}, \hat{\mY}^{L}) = \sum\limits_{i \in \gV^L} \bm{\ell}_{xent} (\vy_i, \hat{\vy}_i )$. Note that GNNs explicitly utilize the graph structure information through learning the mapping from node features and graph adjacency matrix to predicted labels. However, due to the limitations introduced in Sec.~\ref{sec:intro} (inefficiency at inference and poor performance for cold-start nodes), we seek to learn an MLP encoder, i.e., $\bm{H} = f_{\theta} (\bm{X})$ that only takes node features for making predictions. 
\vspace{-5pt}
\subsection{Graph-regularized MLPs}
Graph-Regularized MLPs (GR-MLPs in short) implicitly inject the graph knowledge to the MLP model with an auxiliary regularization term on the node embeddings/predictions over the graph structure~\citep{label-propagation,p-reg,graphmlp}, whose objective function could be generally formulated as: $
   \mathcal{L} = \mathcal{L}_{{sup}} + \lambda \mathcal{L}_{{reg}},  \; \mbox{where} \; \mathcal{L}_{{reg}} = \bm{\ell} (\bm{H}, \bm{A}) \; \mbox{or} \; \bm{\ell} ({\bm{Y}}, \bm{A}) $.
The most representative graph regularization method, Graph Laplacian regularization~\citep{label-propagation, lap-reg},  enforces local smoothness of embeddings/predictions between two connected nodes: $\bm{\ell}(\bm{Y}, \bm{A}) = \textrm{tr}[\bm{Y}^{\top}\bm{L}\bm{Y}]$, where $\bm{L} = \bm{I} - \tilde{\bm{A}} = \bm{I} - \bm{D}^{-1/2}\bm{A}\bm{D}^{-1/2}$ is the (symmetric normalized) Laplacian matrix of the graph. Note that $\bm{Y}$ can be replaced with $\bm{H}$ if one would like to regularize node embeddings instead of predicted labels.

Later works apply advanced forms of regularization, like propagation regularization (P-Reg,~\citet{p-reg}), contrastive regularization~\citep{graphmlp}, etc. Regardless of the minor differences, they are all based on the graph homophily assumption that connected nodes should have similar representations/labels. With the graph structure information implicitly encoded into the model parameters, GR-MLPs can improve the representative power of MLP encoders. However, their performances are still hard to match compared to those of GNN models.
\begin{remark}
    {\rm{(Differences from confusing concepts)}}. Though sound similar, Graph-regularized MLPs(GR-MLPs) are totally different from Graph-augmented MLPs (GA-MLPs). Although trained with implicit graph structure regularization, GR-MLPs make predictions directly through the MLP model. By contrast, GA-MLPs, such as SGC~\citep{sgc}, APPNP~\citep{appnp}, GFNN~\citep{gfnn} and SIGN~\citep{sign} explicitly employ the graph structure to augment the node representation generated from an MLP model. GR-MLP is also different from a recent work named P(ropagational)MLP~\citep{pmlp}. Note that PMLP uses message passing (or graph structure information in general) in testing instead of training, while GR-MLPs use message passing in training instead of testing.
\end{remark}

\subsection{Dimensional Collapse}
Dimensional collapse (also known as spectral collapse in some work~\citep{spectral-collapse}) is a phenomenon in representation learning where the embedding space is dominated by the largest a few singular values (other singular values decay significantly as the training step increases)~\citep{gap_whiten, feature-decorrelation, understand-dimensional-collapse, gnn-express}.  As the actual embedding dimension is usually large, the dimensional collapse phenomenon prevents different dimensions from learning diverse information, limiting their representation power and ability to be linearly discriminated. ~\citet{understand-dimensional-collapse} has analyzed the dimensional collapse phenomenon from a theoretical perspective and attributed it to the effect of strong data augmentation and implicit regularization effect of neural networks~\citep{implicit-reg, gradient-descent-align}. Previous methods usually adopt whitening operation~\citep{feature-decorrelation, w-mse} to mitigate this issue, while such explicit whitening methods are usually computationally inefficient and thus are not applicable to GR-MLP where efficiency is much more important. In this paper, we demonstrate that node embeddings learned from conventional Graph-Regularized MLPs also suffer from dimensional collapse as well. We provide theoretical analysis on how it is caused and develop a computationally efficient soft regularization term to mitigate it.

\section{Dimensional Collapse in GR-MLPs}\label{sec:dimensional-collapse}

In this section, we investigate the reasons behind the weak representation power of previous GR-MLPs. In short, we find that the expressive power of traditional GR-MLPs (e.g., with graph Laplacian regularization~\citep{lap-reg}) is restricted by the dimensional collapse issue, which indicates that the embedding space is dominated by the largest few eigenvalues. We first provide empirical results to demonstrate the existence of the dimensional collapse phenomenon in MLPs with Laplacian regularization. Then we  analyze the causes of it from a theoretical perspective by analyzing the dynamics of Laplacian regularization.

Note that the objective function of Graph Laplacian regularization for semi-supervised node classification tasks could be formulated as follows:
\begin{equation}\label{eqn:lap-reg}
\begin{split}
    \mathcal{L}  & = \bm{\ell}_{xent}(\mY^{L}, \hat{\mY}^{L})  +  \lambda \text{tr} [{\mH}^{\top}{\mL}{\mH}] 
\end{split}
\end{equation}

Following ~\citet{gap_whiten}, we study the eigenvalues of node embeddings' correlation matrix $\mC = \{C_{kk
'}\}\in \sR^{D \times D}$, where $C_{kk'}$ is defined as:
\begin{equation}\label{eqn-corrleation}
    C_{kk'} = \frac{\Sigma_{kk'}}{\sqrt{\Sigma_{kk}\Sigma_{k'k'}}}, \; \mbox{and} \; \bm{\Sigma} = \sum_{i \in \gV}\frac{(\vh_i - \overline{\vh})^{\top}(\vh_i - \overline{\vh})}{|\gV|}
\end{equation}

Note that $\vh_i \in \mathbb{R}^{1\times D}$, $\overline{\vh} =  \sum_{i=1}^{|\gV|} \vh_i / |\gV|$ is the averaged node embedding vector, so $\bm{\Sigma}$ is the covariance matrix of $
\mH$, and we denote $\mC$'s eigenvalues in a descending order by $\{\lambda_1^{\bm{\mC}}, \lambda_2^{\bm{\mC}}, \cdots, \lambda_D^{\bm{\mC}}\}$.

% Note that $\overline{\vh} =  \sum_{i=1}^{|\gV|} \vh_i / |\gV|$ is the average node embedding vector, so $\bm{\Sigma}$ is the covariance matrix of $
% \mH$, and we denote $\mC$'s eigenvalues in a descending order by $\{\lambda_1^{\bm{\mC}}, \lambda_2^{\bm{\mC}}, \cdots, \lambda_D^{\bm{\mC}}\}$. 
\begin{figure*}[t]
    \begin{minipage}[h]{0.65\linewidth}
    \vspace{0pt}
    \centering
\includegraphics[width=1.0\textwidth,angle=0]{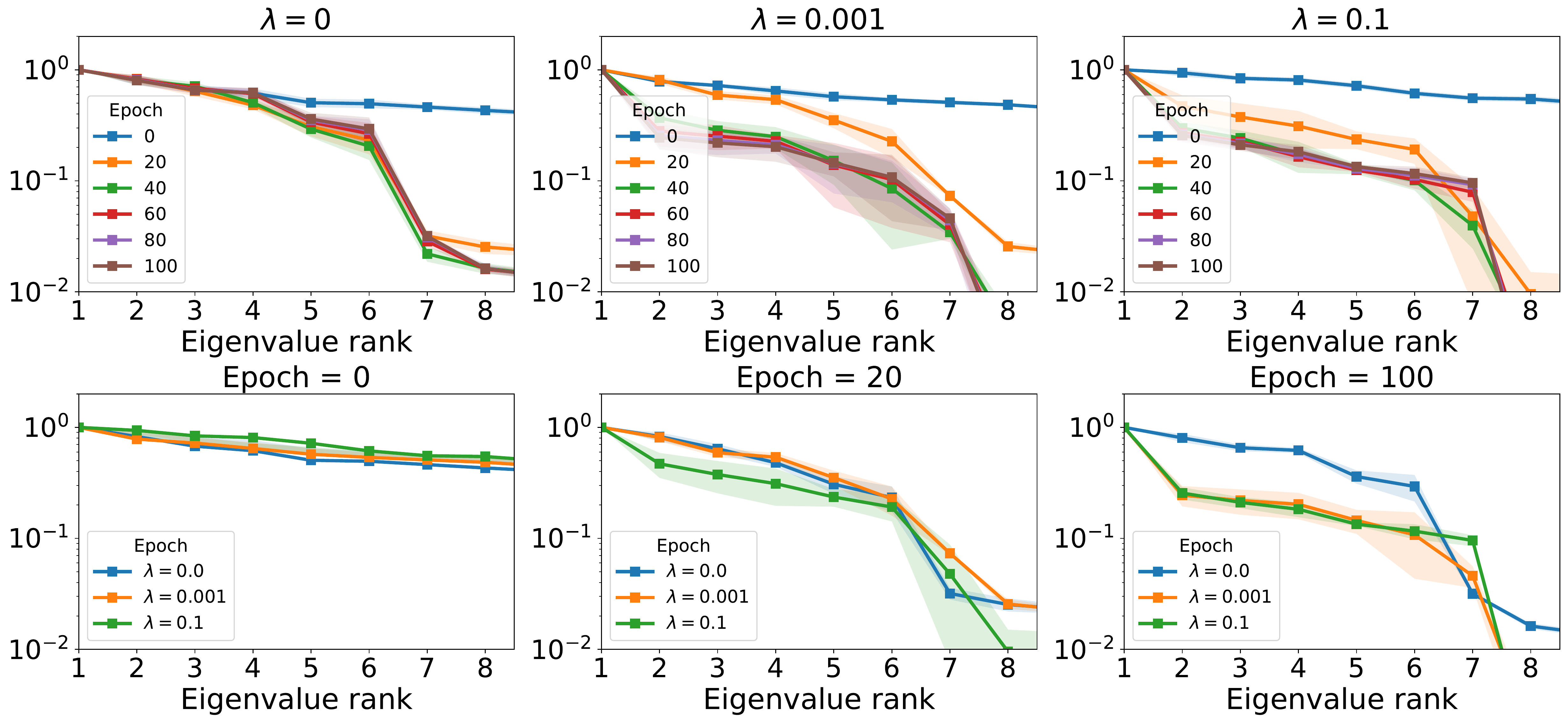}
    \caption{Eigenspectra for node embeddings with different strengths of Laplacian regularization $\lambda$ (the upper three figures), at different training epochs (the lower three figures). x-axis represents the index of sorted eigenvalues and y-axis is the normalized eigenvalue (the ratio to the largest one). The results are averaged over 10 random initialization with $95\%$ confidence intervals.
    }
    \label{fig:eigenvalue-lapreg}
    \end{minipage}
    \begin{minipage}[h]{0.3\linewidth}
    \vspace{0pt}
    \centering
\includegraphics[width=1.0\textwidth,angle=0]{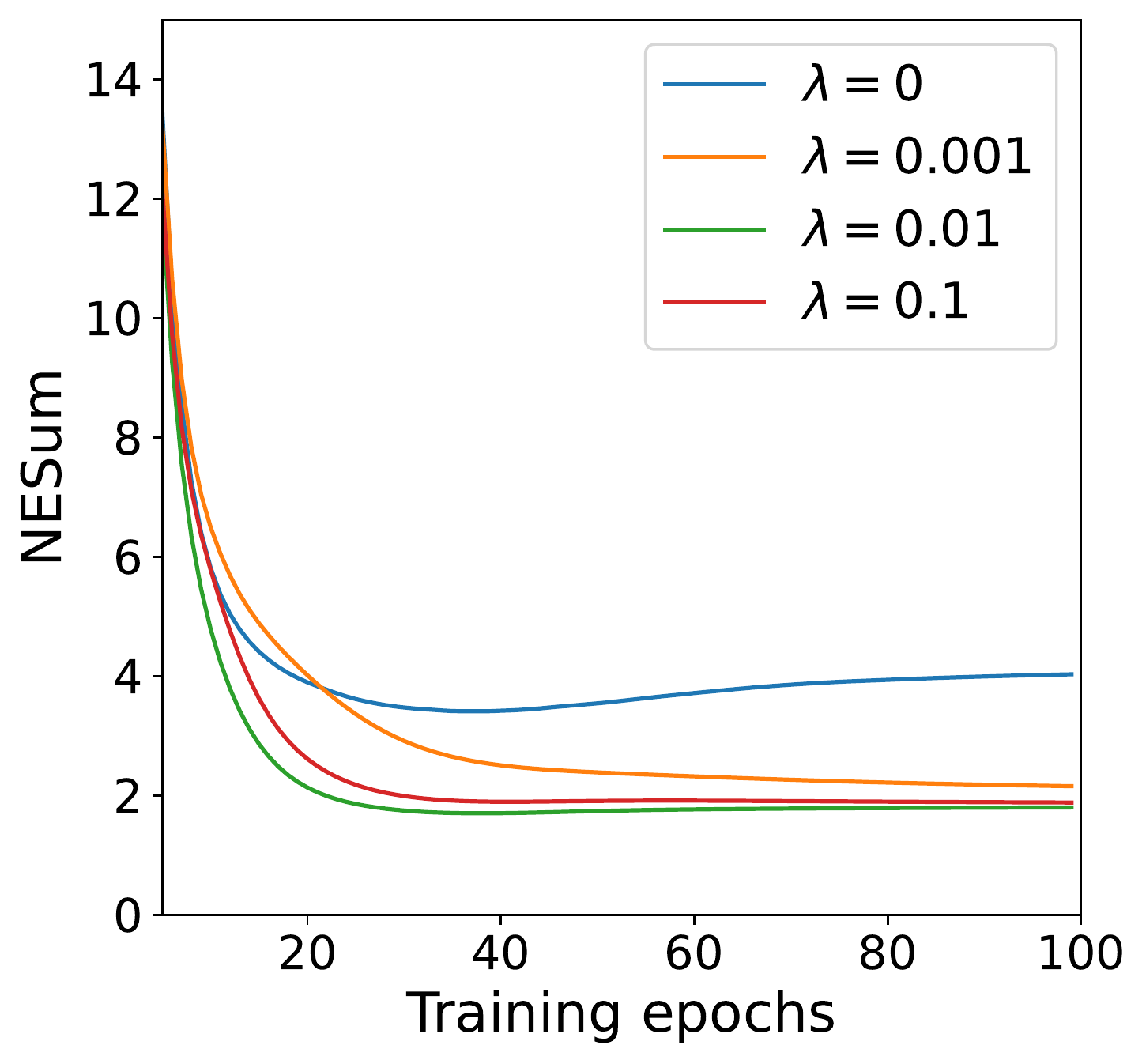}
    \caption{Evolving of NESum as the training epoch increases, with different regularization factors.}
    \label{fig:eigensum}
    \end{minipage}
\end{figure*}
\paragraph{Empirical Observations.}
We train a 3-layer MLP model using Eq.~\ref{eqn:lap-reg} as the objective function on \cora dataset. To study the relationship between Laplacian regularization and the dimensional collapse phenomenon, we try different  regularization factor $\lambda$ (i.e., $0$, $0.001$, $0.1$ respectively). Note that $\lambda = 0$ corresponds to a pure MLP without regularization. Fig~\ref{fig:eigenvalue-lapreg} plots the evolvement of top eigenvalues (which are re-normalized as the ratio between each eigenvalue to the largest one $\lambda_i^{\bm{\mC}} / \lambda_1^{\bm{\mC}}$) as the training step increases and with different factors $\lambda$. We can observe that without Laplacian regularization (i.e., $\lambda = 0$), the decay of top eigenvalues is very slow and is almost negligible (e.g., $\lambda^{\bm{\mC}}_6 / \lambda^{\bm{\mC}}_1 \ge 0.5$ even after $100$ steps). By contrast, even with a small regularization factor $\lambda^{\bm{\mC}} = 0.001$, we can observe a fast decay rate (e.g, $\lambda^{\bm{\mC}}_5 / \lambda^{\bm{\mC}}_1 \le 0.2 $ after $40$ steps). This phenomenon is even more remarkable with a large factor. These observations demonstrate a positive correlation between Laplacian regularization and the dimensional collapse phenomenon.

We further employ \textit{normalized eigenvalue sum} (NESum) introduced in~\citet{gap_whiten} as a metric to measure the extent of the dimensional collapse. Formally, NESum is defined as the ratio between the summation of all eigenvalues and the largest one: $ {\rm NESum}(\{\lambda_i^{\mC}\}) \triangleq \sum_{i=1}^{D} {\lambda_i^{\mC}} /{\lambda_1^{\mC}}$. Intuitively, a large NESum value indicates that the eigenvalues are fluently distributed, while a very small one indicates the dimensional collapse phenomenon (the largest eigenvalue becomes dominant).

In Fig.~\ref{fig:eigensum}, we plot the evolution of NESum with different regularization strengths. It is observed that 1) NESum decreases as training goes on because the model learns to pay more attention to important features for downstream classification tasks. 2) NESum trained with purely cross-entropy loss converges to a high value. 3) With additional Laplacian regularization, NESum decreases quickly and converges to a small value even if the regularization factor $\lambda$ is small. The above observations demonstrate that Laplacian regularization leads to a larger decay rate of top eigenvalues. The significant decay rate will make the learned representations less informative for classifications due to \textit{information loss}~\citep{gnn-express}.

\paragraph{Theoretical Analysis.} The empirical results above have verified the inner connection between Laplacian regularization and dimensional collapse. We'd like to further study how the optimization of Laplacian regularization leads to dimensional collapse. 

To simplify the analysis, we first consider a simple linear model as the encoder to learn node embeddings, i.e., $\mH = \mX\mW$, where $\mW \in \sR^{F\times D}$ is the weight matrix (we further assume $F = D$ in this part for simplicity). The model (i.e., the weight matrix $\mW$) is optimized using stochastic gradient descent. Then we have the following lemma on the evolvement of the weight matrix's singular values.
\begin{lemma}\label{lemma:weight_matrix-shrink}
{\rm{(Shrinking singular-space of weight matrix.)}} Consider the linear model above which is optimized with $\mathcal{L}_{reg} =  \text{tr} [{\mH}^{\top}{\mL}{\mH}]$. Let $\mP = \mX^{\top} \mL \mX = \sum\limits_{ij}L_{ij}\vx_i \cdot \vx_j^{\top}$ and denote its non-ascending eigenvalues by $\{\lambda^{\mP}_1, \lambda^{\mP}_2, \cdots, \lambda^{\mP}_D \}$. Denote the randomly initialized weight matrix by $\mW(0)$ and the updated weight matrix at time $t$ by $\mW(t)$, respectively. We further denote the non-ascending singular values of $\mW$ at time $t$ by $\{\sigma^{\mW}_i (t)\}_{i=1}^D$. Then the relative value of the smaller eigenvalues to the larger ones will decrease as $t$ increases. Formally,
% \begin{equation}
    $\frac{\sigma^{\mW}_i(t)}{\sigma^{\mW}_j(t)} \le \frac{\sigma^{\mW}_i(t')}{\sigma^{\mW}_j(t')}, \;   \; \forall \; t < t', i\le j. $
% \end{equation}
Furthermore, if the following condition holds: $\lambda^{\mP}_1 \ge \cdots \ge \lambda^{\mP}_d > \lambda^{\mP}_{d+1} \ge \cdots \ge \lambda^{\mP}_D $, then
\begin{equation}
    \lim_{t \rightarrow \infty} \frac{\sigma^{\mW}_i(t)}{\sigma^{\mW}_j(t)} = 0, \; \forall \; i \le d \; \mbox{and} \; j \ge d+1.
\end{equation}
\end{lemma}
See proof in Appendix~\ref{proof:weight_matrix-shrink}. Lemma~\ref{lemma:weight_matrix-shrink} indicates that the singular values of $\mW$ (in proportional to the larger ones) shrink as the training step increases. With Lemma~\ref{lemma:weight_matrix-shrink}, we can conclude the following theorem that reveals a dimensional collapse phenomenon under this condition:
\begin{theorem}\label{theorem:dimensional-collapse}
{\rm (Laplacian regularization leads to dimensional collapse.)} For the linear model  above optimized with Graph Laplacian regularization, the embedding space of nodes tends to be dominated by the largest a few eigenvalues. Specifically, if the covariance matrix of input features is an identity matrix, we have:
\begin{equation}
    \lim_{t \rightarrow \infty} \frac{\lambda^{\mC}_i(t)}{\lambda^{\mC}_j(t)} = 0, \; \forall \; i \le d \; \mbox{and} \; j \ge d+1.
\end{equation}
\end{theorem}
See proof in Appendix~\ref{proof:dimensional-collapse}. Theorem~\ref{theorem:dimensional-collapse} reveals that with the effect of Graph Laplacian regularization, the eigenspectrum is dominated by its largest few eigenvalues, leading to the dimensional collapse phenomenon. 

In a more general case, the encoder should be more complicated (e.g., an MLP with non-linearity) rather than a linear model. In this case, we can study the asymptotic behavior (e.g., dynamics) in feature space. Gradient descent with step size $\tau$ derives the following update rule of the node embedding matrix:
\begin{equation}
    \mH^{(t+1)} = [(1-2\tau)\mI + 2\tau \tilde{\mA}]\mH^{(t)}.
\end{equation}
Let $\tau = 1/2$, we have a step-wise updating formula $\mH^{(t+1)} = \tilde{\mA}\mH^{(t)}$, where $\tilde{\mA} = \mD^{-1/2}\mA\mD^{-1/2}$. \citet{gnn-express} has proved that such an updating rule leads to a shrinking low-dimensional embedding subspace as $t \rightarrow \infty$, which restricts the expressive power due to information loss.

\section{Overcoming Dimensional Collapse via Orthogonality Regularization}
\subsection{Explicit Regularization on the Correlation Matrix}

Our thorough analysis in Sec.~\ref{sec:dimensional-collapse} reveals that the poor performance of GR-MLPs could be attributed to less-expressive node representations (due to dimensional collapse). Specifically, we establish that the eigenspectrum of the embeddings' correlation matrix is dominated by the largest eigenvalue (different dimensions are \textbf{over-correlated}).

In contrast to dimensional collapse, whitened representations have an identity correlation matrix with equally distributed eigenvalues. Motivated by this, a natural idea should be enforcing a soft regularization term on the correlation matrix of node embeddings, e.g., minimizing the distance between $\mC$ and the identity matrix $\mI$:
\begin{equation}\label{eqn:corr_reg}
    \bm{\ell}_{corr\_reg} = \Vert \bm{C} - \bm{I} \Vert_F^2 = \sum\limits_{i=1}^{d} (1-{{C}}_{ii})^2 + \sum\limits_{i \neq j} {C}_{ij}^2 = \sum\limits_{i \neq j} {C}_{ij}^2 .
\end{equation}
Note that the on-diagonal terms ${C}_{ii} = 1$ for all $i$, so Eq.~\ref{eqn:corr_reg} is essentially forcing the off-diagonal terms of the correlation matrix to become zero, or in other words, making the embeddings \textbf{orthogonal}, so that different dimensions of node embeddings can capture orthogonal information. 
One may directly equip Eq.~\ref{eqn:corr_reg} with existing GR-MLPs for alleviating the dimensional collapse issue. However, we would like to design a more general, flexible, and elegant formulation that can handle high-order connectivity and non-homophily graphs~\citep{geom-gcn,heterophily-survey}.
We then introduce \modelname, a powerful and flexible GR-MLP model, step by step.
\subsection{Graph-Regularized MLP with \modelname}\label{sec:method-ortho-reg}
% The framework of the proposed method is illustrated below:
Similar to previous GR-MLPs, we first use an MLP encoder to map raw node features to the embeddings. This process can be formulated as $\mH = {\rm MLP}_{\theta}(\mX)$, where $\mX = \{\vx_i\}_{i=1}^{|\gV|}$ is raw node features while $\bm{H} = \{ \vh_i\}_{i=1}^{|\gV|}$ is the embedding matrix.

% \textbf{Encoding.} The encoder is an MLP model which takes merely node features as input. The embeddings $\bm{H} \in \mathbb{R}^{N \times d}$ could be formulated as $\bm{H} = \textrm{MLP}_{\theta}(\bm{X})$, where $\bm{X} \in \mathbb{R}^{N \times F}$ is the input node features.
The next question is what kind of graph structure information is more beneficial. Previous GR-MLPs resort to either edge-centric smoothing~\citep{label-propagation,lap-reg} or node-centric matching~\citep{p-reg, graphmlp}. While recent studies indicate that the node-centric method is more appropriate for node-level tasks as edge-centric methods overemphasize the ability to recover the graph structure~\citep{p-reg}. Inspired by this, we employ a \textbf{neighborhood abstraction} operation to summarize the neighborhood information as guidance of the central node. Formally, for a node $i \in \gV$ and the embeddings of its (up-to) $T$-hop neighbors $\{\bm{h}_j\}^{(1:T)}(i)$, we can get the summary if its $T$-hop neighborhoods through a pooling function $\vs_i = {\rm Pool}( \{\bm{h}_j\}^{(1:T)}(i) )$. The exact formulation of the pooling function could be flexible to fit graphs with different properties. However, here we consider a simple average pooling of node embeddings from different order's neighborhoods for simplicity, which can work in most cases:
\begin{equation}\label{eqn:neighbor-summary}
    \mS = \sum_{t=1}^T \tilde{\mA}^t\mH / L, \; \mbox{where} \; \tilde{\mA} = \mA {\mD}^{-1}.
\end{equation}

To make the node embeddings aware of structural information, we employ the following regularization term on the cross-correlation matrix of node embeddings $\mH$ and summary embeddings $\mS$:
\begin{equation}\label{eqn:ortho-reg}
    \gL_{reg} = -\alpha \sum\limits_{k=1}^D C_{kk} + \beta\sum\limits_{k\neq k'}C_{kk'}^2,
\end{equation}
where $\mC = \{C_{kk'}\} \in \sR^{D\times D}$ is the cross-correlation matrix of $\mH$ and $\mS$. We show in the following theorem that with Eq.~\ref{eqn:ortho-reg}, the node embeddings will be locally smoothed and at the same time, prevent dimensional collapse:
\begin{theorem}\label{theorem:ortho-reg}
    Assume $T$ = 1 and $\mH$ are free vectors. Let $\mH^*$ be a global optimizer of Eq.~\ref{eqn:ortho-reg}, then $\mH^*$ is smoothed over the graph structure and is orthogonal.
\end{theorem}
See proof in Appendix~\ref{proof:ortho-reg}. Finally, we can employ an additional linear layer to make predictions $\mY = {\rm LIN}_{\phi}(\mH)$. Then the final objective function to be optimized is:
\begin{equation}
    \gL = \bm{\ell}_{xent}(\mY^L, \hat{\mY}^L) - \alpha \sum\limits_{k=1}^D C_{kk} + \beta\sum\limits_{k\neq k'}C_{kk'}^2,
\end{equation}
where $\alpha, \beta$ are trade-off hyperparameters to balance the strengths of regularization.
\begin{remark}
     With a well-trained model, we can mkae prediction for an upcoming node with feature $\bm{x}$ with $\bm{y} = {\rm {Lin}}_{\phi} ({\rm {MLP}}_{\theta}(\bm{x}))$ quickly, and without the help of graph structure.
\end{remark}

\section{Experiments}\label{sec:experiments}
In this section, we conduct experiments to evaluate \modelname by answering the following research questions:
\begin{itemize}
    \item \textbf{RQ1}: What's the performance of \modelname on common transductive node classification tasks compared with GNN models and other MLP models? (Sec.~\ref{exp:transductive})
    \item \textbf{RQ2}: On cold-start settings where we do not know the connections of testing nodes, can \modelname demonstrate better performance than other methods? (Sec.~\ref{exp:inductive})
    \item \textbf{RQ3}: Does \modelname mitigate the dimensional collapse issue, and is each design of \modelname really necessary to its success? (Sec.~\ref{exp:abl})
    \item \textbf{RQ4}: Can \modelname demonstrate better robustness against structural perturbations compared with Graph Neural Networks? (Sec.~\ref{sec:exp-robustness})
\end{itemize}
Due to space limits, we defer the experiments on heterophily graphs and scalability comparison in Appendix~\ref{appendix-exp-hete} and Appendix~\ref{appendix-exp-scale}, respectively. A brief introduction of the baselines is given in Appendix~\ref{baselines}.

\subsection{Experiment setups}\label{sec:exp-steups}
\textbf{Datasets.} We consider $7$ benchmark graph datasets and their variants in this section: \cora, \citeseer, \pubmed, \computer, \photo, \cs, and \physics as they are representative datasets used for semi-supervised node classification~\citep{gcn, graphmlp, glnn, coldbrew}. The detailed introduction and statistics of them are presented in Appendix~\ref{appendix-exp-detail}. To evaluate \modelname on large-scale graphs, we further consider two OGB datasets~\citep{ogb}: \arxiv and \products. Note that the two OGB datasets are designed for fully-supervised node classification tasks, so we defer their results to Appendix~\ref{appendix-exp-add}. 

\textbf{Implementations.} If not specified, we use a two-layer MLP model as the encoder to generate node embeddings, then another linear layer takes node embeddings as input and outputs predicted node labels. We use Pytorch to implement the model and DGL~\citep{dgl} to implement the neighborhood summarizing operation in Eq.~\ref{eqn:neighbor-summary}. If not specified, all our experiments are conducted on an NVIDIA V100 GPU with 16G memory with Adam optimizer~\citep{adam}.
\subsection{Transductive Semi-supervised Node Classification (RQ1)}\label{exp:transductive}
\begin{table*}[t] 
    \centering
    \caption{Prediction accuracy of semi-supervised node classification tasks on the seven benchmark graphs. \modelname outperforms powerful GNN models and competitive MLP-architectured baselines on 6 out of 7 datasets.}
	\label{tbl-exp-semi}
	\small
	\begin{threeparttable}
        {
        \scalebox{0.87}
        {
		\begin{tabular}{c|l|ccccccc}
			\toprule[0.8pt]
		 & Methods & \cora & \citeseer  & \pubmed & \texttt{Computer} & \texttt{Photo} & \texttt{CS} & \texttt{Physics} \\
			\midrule
		\multirow{3}{*}{GNNs} & SGC & 81.0$\pm$0.5 & 
            71.9$\pm$0.5 & 78.9$\pm$0.4 & 80.6$\pm$1.9 & 90.3$\pm$0.8 & 87.9$\pm$0.7 & 90.3$\pm$1.4 \\
          & GCN &  82.2$\pm$0.5 & 71.6$\pm$0.4 & 79.3$\pm$0.3 &         82.9$\pm$2.1 & 91.8$\pm$0.6 & 89.9$\pm$0.7 & 91.9$\pm$1.2 \\
	   &  GAT & 83.0$\pm$0.7 & 72.5$\pm$0.7 & 79.0$\pm$0.3 & 82.5$\pm$1.6 & 91.4$\pm$0.8 & 90.5$\pm$0.8 & 92.3$\pm$1.5  \\
      	\midrule[0.5pt]
        KD-MLPs & GLNN &  82.6$\pm$0.5 & 72.8$\pm$0.4 & 80.2$\pm$0.6 & 82.1$\pm$1.9 & 91.3$\pm$1.0 & 92.6$\pm$1.0 & \textbf{93.3}${\bm{\pm}}$\textbf{0.5} \\
          \midrule[0.5pt]
         \multirow{5}{*}{GR-MLPs} & MLP & 59.7$\pm$1.0 & 57.1$\pm$0.5 & 68.4$\pm$0.5 & 62.6$\pm$1.8 & 76.2$\pm$1.4 & 86.9$\pm$1.0 & 89.4$\pm$0.7 \\
  	 & Lap-Reg & 60.3$\pm$2.5 & 58.6$\pm$2.4 & 68.7$\pm$1.4 & 62.6$\pm$2.0 & 76.4$\pm$1.1 & 87.9$\pm$0.6 & 89.5$\pm$0.5 \\
  	 & P-Reg & 64.4$\pm$4.5 & 61.1$\pm$2.1 & 72.3$\pm$1.7 & 68.9$\pm$3.3 & 79.7$\pm$3.7 & 90.9$\pm$1.9 & 91.6$\pm$0.7 \\
  	 & GraphMLP & 79.5$\pm$0.6 & 73.1$\pm$0.4 & 79.7$\pm$0.4 & 79.3$\pm$1.7 & 90.1$\pm$0.5 & 90.3$\pm$0.6 & 91.6$\pm$0.8 \\
          & N2N & 83.2$\pm$0.4 & 73.3$\pm$0.5 & 80.9$\pm$0.4 & 81.4$\pm$1.6 & 90.9$\pm$0.7 & 91.5$\pm$0.7 & 91.8$\pm$0.7 \\
                \midrule[0.5pt]
  	Ours &  \modelname &  \textbf{84.7}${\bm{\pm}}$\textbf{0.4} & \textbf{73.5}${\bm{\pm}}$\textbf{0.4} & \textbf{82.8}${\bm{\pm}}$\textbf{0.5} & \textbf{83.7}${\bm{\pm}}$\textbf{1.5} & \textbf{92.3}${\bm{\pm}}$\textbf{1.0} & \textbf{92.9}${\bm{\pm}}$\textbf{1.1} & 92.8$\pm$0.8  \\
			\bottomrule[0.8pt] 
		\end{tabular}
  }
  }
	\end{threeparttable}
\end{table*}
We first evaluate our method on transductive semi-supervised node classification tasks. For comparison, we consider three types of baseline models: 1) Graph Neural Networks (GNNs), including SGC~\citep{sgc}, GCN~\citep{gcn} and GAT~\citep{gat}. 2) Representative knowledge distillation (KD-MLP) method GLNN~\citep{glnn}. 3) Basic MLP and GR-MLP models, including Laplacian regularization (Lap-Reg,~\citet{label-propagation}, \citet{lap-reg}), Propagation Regularization (P-Reg,~\citet{p-reg}), GraphMLP~\citep{graphmlp}, and Node-to-Neighborhood Mutual Information Maximization (N2N,~\citet{n2n})

For each dataset, we use $20$ nodes per class for training, $500$ nodes for validation, and another $1000$ nodes for testing. For \cora, \citeseer, and \pubmed we use the public split, while for the remaining datasets, we split randomly. We report the average prediction accuracy with standard deviation over 20 random trials in Table~\ref{tbl-exp-semi}.

As demonstrated in the table, \modelname outperforms previous GR-MLPs by a large margin, which greatly validates the importance and effectiveness of orthogonal node embeddings. Compared with the competitive knowledge distillation method GLNN, \modelname also demonstrates better performance on $6$ out of $7$ graphs. It is also worth noting that our method even outperforms powerful GNN models such as GCN and GAT, which indicates that node features of the graphs are less exploited by these GNN models. In contrast, our method can fully exploit the potential of node features.

\subsection{Inductive Node Classification for Cold-Start Scenarios (RQ2)}\label{exp:inductive}
To evaluate the performance of \modelname on cold-start scenarios where the connections between newly encountered nodes and existing nodes are missing, we follow the setups in ColdBrew that selects a proportion of nodes as \textit{isolated} nodes which will be removed from the original graph. Then the model is evaluated on the isolated nodes in the testing set. Due to the space limit, we present the detailed setups and evaluation methods in Appendix~\ref{appendix-exp-inductive}. Besides the baselines used in~\citet{coldbrew}, we also include GLNN for a fair comparison.
% To evaluate the performance of \modelname on cold-start scenarios where the connections between newly encountered nodes and existing nodes are missing or few, we follow the setups in ColdBrew~\citep{coldbrew}, another KD-MLP method but specially designed for cold-start settings, and conduct experiments on the three citation networks \cora, \citeseer and \pubmed. Specifically, for each dataset, we first rank among the nodes according to their degrees. Then the bottom 10\% nodes with the lowest degree are removed from the graph and are termed as \textit{isolated} nodes. The testing isolated nodes are invisible during the whole training phase. We also leave out the remaining bottom 10\% nodes with small degrees as \textit{tail} nodes to study the generalization ability of \modelname when the connectivity of encountered nodes is partially known. In consistency with the evaluation method in~\citet{coldbrew}, we use the fixed 20 nodes per class for training and all the remaining nodes for testing. Besides the baselines used in~\citet{coldbrew}, we also include GLNN.
\begin{table}[t] 
    \centering
    \caption{Test accuracy on the isolated nodes.}
	\label{tbl-exp-cold}
	\small
        {   
    	\begin{threeparttable}
        {
        \scalebox{0.9}{
	\begin{tabular}{llcccccc}
	\toprule[0.8pt]
		 \multicolumn{2}{c}{{Methods}} & \cora & \citeseer  & \pubmed  \\
	\midrule
		 \multirow{2}{*}{{GNNs}} & GCN  & 53.02$\pm$1.78 & 47.09$\pm$1.38 & 71.50$\pm$2.21  \\
           & GraphSAGE & 55.38$\pm$1.92 & 41.46$\pm$1.57 & 69.87$\pm$2.13   \\
          \specialrule{0em}{1pt}{1pt}
          \cline{1-2}
          \specialrule{0em}{1pt}{1pt}
        \multirow{2}{*}{{KD-MLPs}} & ColdBrew & 58.75$\pm$2.11 & 53.17$\pm$1.41 & {72.31$\pm$1.99}   \\
        & GLNN & {59.34$\pm$1.97} & {53.64$\pm$1.51} & 73.19$\pm$2.31  \\
        \specialrule{0em}{1pt}{1pt}
          \cline{1-2}
          \specialrule{0em}{1pt}{1pt}
            \multirow{4}{*}{{GR-MLPs}} & MLP & 52.35$\pm$1.83 & 53.26$\pm$1.41 & 65.84$\pm$2.08  \\
             & GraphMLP & 59.32$\pm$1.81 & 53.17$\pm$1.48 & 72.33$\pm$2.11  \\
             & \modelname & {\textbf{61.93}${\bm{\pm}}$\textbf{1.77}} & {\textbf{56.31}${\bm{\pm}}$\textbf{1.54}} & {\textbf{73.42}${\bm{\pm}}$\textbf{1.99}}  \\
        \bottomrule[0.8pt] 
	\end{tabular}}}
	\end{threeparttable}}
	
\end{table}

% \begin{table}[h] 
% 	\centering
% 	\caption{Test accuracy on the isolated nodes.}
% 	\label{tbl-exp-cold}
% 	\small
%    {
% 	\begin{threeparttable}
%         {
%         \scalebox{0.8}{
% 	\begin{tabular}{llcccccc}
% 	\toprule[0.8pt]
% 		 \multicolumn{2}{c}{{Methods}} & \cora & \citeseer  & \pubmed  \\
% 	\midrule
% 		 \multirow{2}{*}{{GNNs}} & GCN  & 53.02$\pm$1.78 & 47.09$\pm$1.38 & 71.50$\pm$2.21  \\
%            & GraphSAGE & 55.38$\pm$1.92 & 41.46$\pm$1.57 & 69.87$\pm$2.13   \\
%           \specialrule{0em}{1pt}{1pt}
%           \cline{1-2}
%           \specialrule{0em}{1pt}{1pt}
%         \multirow{2}{*}{{KD-MLPs}} & ColdBrew & 58.75$\pm$2.11 & 53.17$\pm$1.41 & {72.31$\pm$1.99}   \\
%         & GLNN & {59.34$\pm$1.97} & {53.64$\pm$1.51} & 73.19$\pm$2.31  \\
%         \specialrule{0em}{1pt}{1pt}
%           \cline{1-2}
%           \specialrule{0em}{1pt}{1pt}
%             \multirow{4}{*}{{GR-MLPs}} & MLP & 52.35$\pm$1.83 & 53.26$\pm$1.41 & 65.84$\pm$2.08  \\
%              & GraphMLP & 59.32$\pm$1.81 & 53.17$\pm$1.48 & 72.33$\pm$2.11  \\
%              & \modelname (Ours) & {\textbf{61.93}${\bm{\pm}}$\textbf{1.77}} & {\textbf{56.31}${\bm{\pm}}$\textbf{1.54}} & {\textbf{73.42}${\bm{\pm}}$\textbf{1.99}}  \\
%         \bottomrule[0.8pt] 
% 	\end{tabular}}}
% 	\end{threeparttable}}
% \end{table}

In Table.~\ref{tbl-exp-cold}, we report the experimental results of \modelname and baseline methods on the isolation nodes. As demonstrated in the table, for isolated nodes whose connectivity in the graph is unknown, GNN models perform poorly as they require both the node features and graph structure for accurate inference. By contrast, MLP-based models generalize better on isolated nodes as they make the best of the available node features. The proposed \modelname outperforms both GNNs and MLPs (including KD MLPs and GR-MLPs) baselines. 

\subsection{Studies of \modelname (RQ3)}\label{exp:abl}
\subsubsection{Does \modelname mitigate dimensional collapse?}
In Sec.~\ref{sec:dimensional-collapse} we have attributed the limitation of previous GR-MLPs to the dimensional collapse phenomenon, and in Sec.~\ref{sec:method-ortho-reg} we have proposed \modelname to mitigate such a problem from a theoretical perspective. In this part, we would like to empirically show that \modelname can avoid the dimensional collapse issue by keeping node embeddings' eigenspectra.

\begin{figure}[t]
    \centering
    \includegraphics[width = 0.95\linewidth]{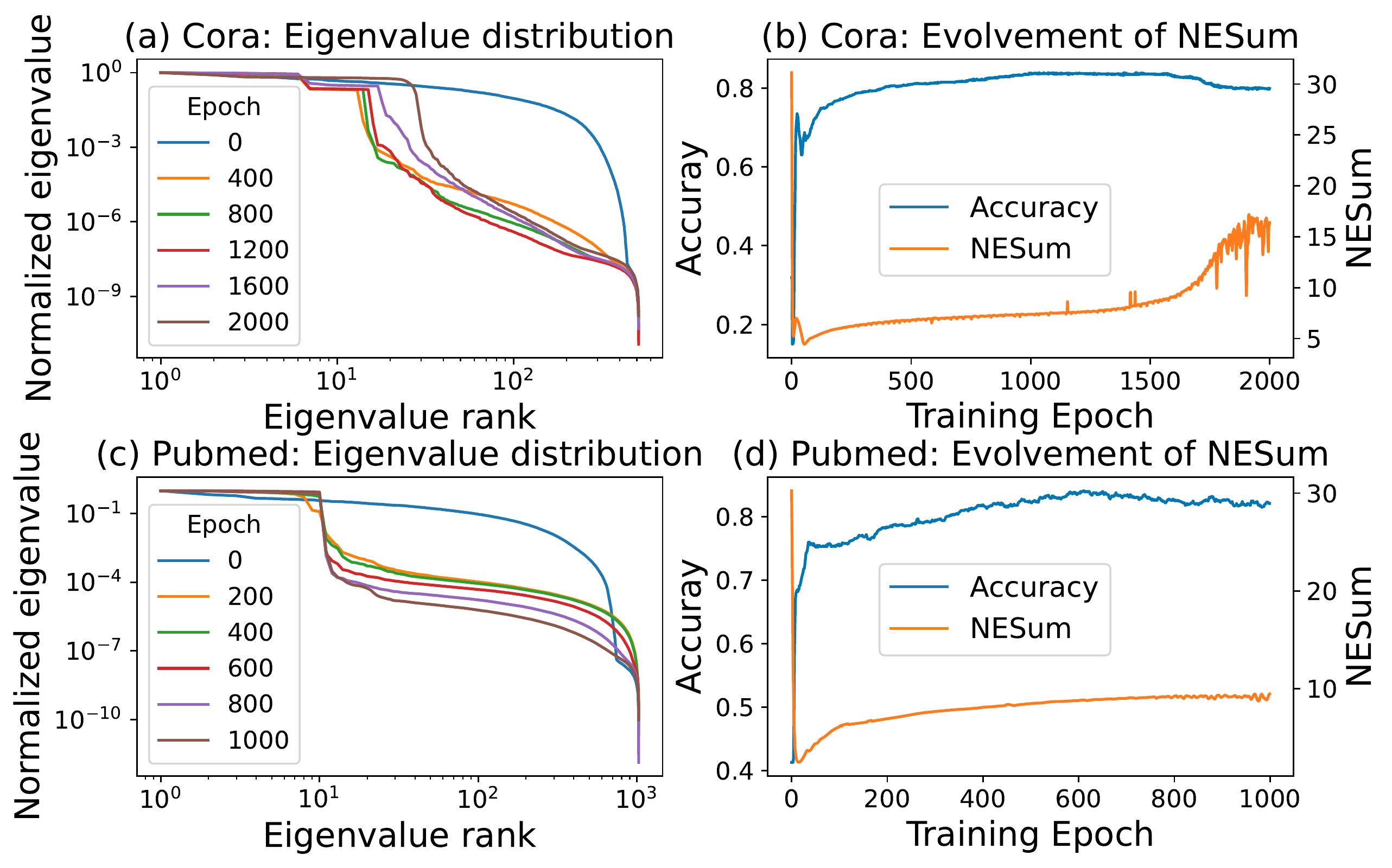}
    \caption{Visualization of \modelname's impact on node embeddings' Eigenspectra on \cora and \pubmed.}
    \label{fig:vis}
\end{figure}

In consistency with the settings in Sec.~\ref{sec:dimensional-collapse}, we evaluate the embeddings learned from \modelname at different training epochs (we take both \cora and \pubmed for illustrations). The decay of eigenvalues of node embeddings' correlation matrix at different epochs is plotted in Fig.~\ref{fig:vis} (a) and (c). It is observed that the top eigenvalues are well-reserved thanks to the explicit regularization of node embeddings' correlation matrix. In Fig.~\ref{fig:vis} (b) and (d) we also plot the change of testing accuracy as well as the NESum value as the training epoch increases, from which we could observe a positive relationship between the NESum value and the test accuracy: neglecting the initial oscillations, we notice the test accuracy will grow smoothly as the NESum value increases and will reach its peak when NESum overwhelms (\cora) or converges (\pubmed).
The above observations demonstrate that \modelname does mitigate the dimensional collapse problem and lead to a more powerful model.
\subsubsection{Ablation Studies}
We then conduct ablation studies to study the effect of different components of \modelname, and we present the results in Table \ref{tbl-exp-abl}. We first study the impact of the two regularization terms by setting the corresponding factors ($\alpha$ and $\beta$) to $0$, respectively. When $\alpha = 0$ (i.e., only decorrelating different dimensions), we observe that the model's performance is even worse than the pure MLP model (see in Table~\ref{tbl-exp-semi}). This indicates that adding orthogonal regularization is not always beneficial (e.g., for vanilla MLP), but is indeed beneficial for GR-MLPs. By contrast, without orthogonal regularization (i.e., $\beta = 0$), the power of structure regularization is restricted, and decorrelating different dimensions can boost performance greatly. We further investigate whether considering a larger neighborhood would improve the model's performance. The empirical results demonstrate that considering a larger neighborhood improves the performance compared to only using first-order neighborhoods, but $T = 2$ is already optimal for most datasets. 
\begin{table}[t] 
    \centering
    \caption{Effects of different components of \modelname}
	\label{tbl-exp-abl}
	\small
    {   
    \begin{threeparttable}
    {
    \scalebox{1.0}{
		\begin{tabular}{c|ccc}
			\toprule[0.8pt]
		    Variants  & {\cora} & {\citeseer} & {\pubmed} \\
		    \midrule[0.5pt]
		    Baseline  & 84.7  & 73.5 & 82.8  \\
		    \midrule[0.5pt]
			$\alpha = 0$  & 54.7 & 51.4 & 47.2  \\
			$\beta = 0$  & 79.3 & 68.7 & 76.8 \\
                \midrule[0.5pt]
                $T = 1$ & 83.9 & 72.9 & 82.1 \\
                $T = 2$ & \textbf{84.7} & \textbf{73.5} & \textbf{82.8}\\
                $T = 3$ & 84.3 & 73.3 & 82.5\\
			\bottomrule[0.8pt]
		\end{tabular}}
        } 
	\end{threeparttable}
	}
\end{table}

\subsubsection{Hyperparameter Analysis}
\begin{figure}[t]
    \centering
    \includegraphics[width = 1.0\linewidth]{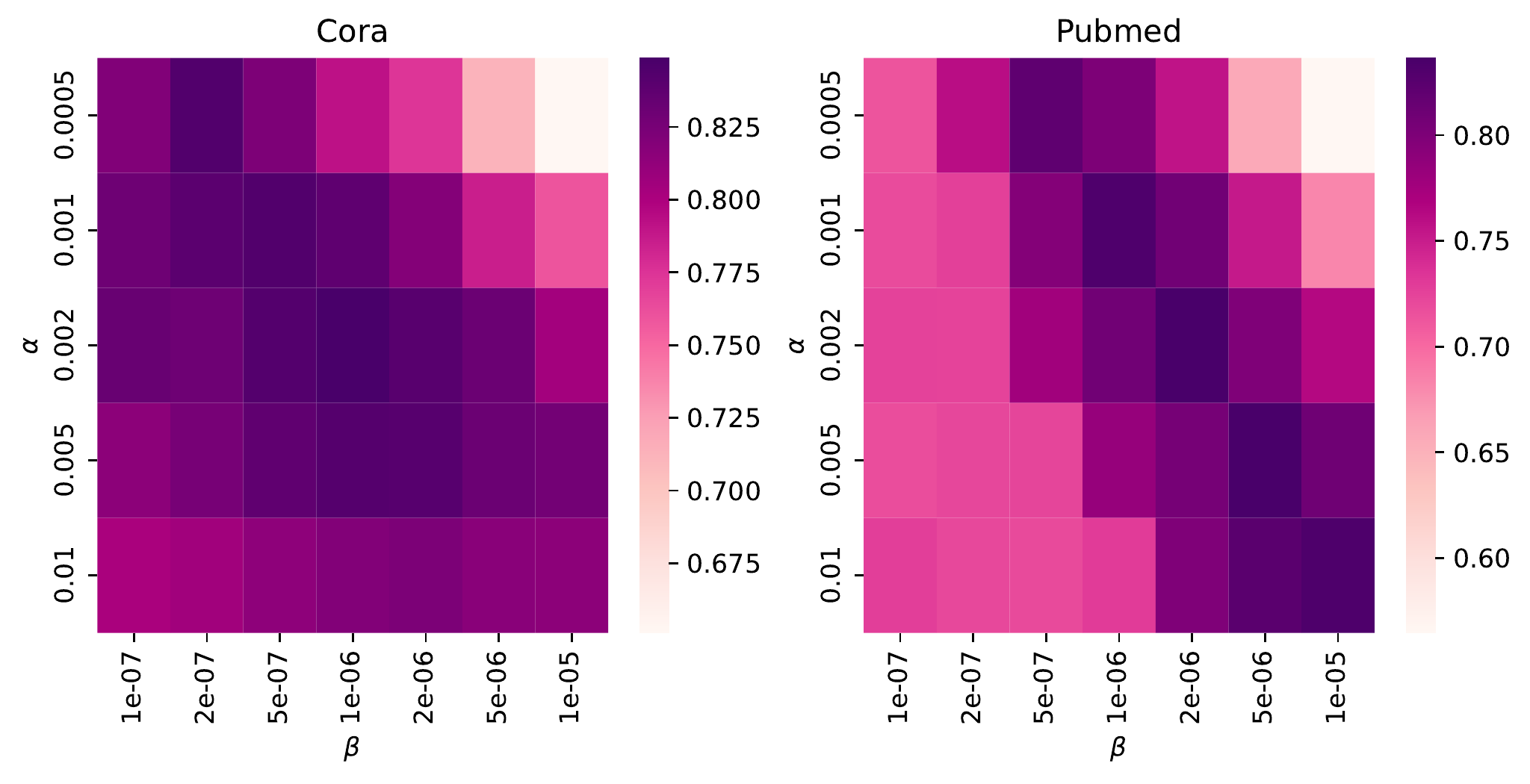}
    \caption{Performance heat map when using different $\alpha$, $\beta$ combinations in Eq.~\ref{eqn:ortho-reg}, on \cora and \pubmed.}
    \label{fig:sense}
\end{figure}
We further study how the two trade-off hyperparameters $\alpha$ and $\beta$ affect the performance of \modelname. We try different combinations of $\alpha$ and $\beta$ on \cora, and \pubmed (we defer the results on \citeseer to Appendix~\ref{appendix-exp-sense} due to space limit), and plot the performance heatmap in Fig.~\ref{fig:sense}. 

The conclusion is very interesting: the performance of \modelname is not very sensitive to a specific value of $\alpha$ or $\beta$. In other words, for a reasonable value of $\alpha$ ($\beta$), we can easily find another value of $\beta$ ($\alpha$) that can achieve similarly high performance. The ratio between $\alpha$ and $\beta$ seems much more important. From Fig.~\ref{fig:sense}, we can observe that for \cora, $\alpha/\beta = 2* 10^3$, and for \pubmed, $\alpha/\beta = 1* 10^3$ can lead to the optimal performance; changing the value of $\alpha$ while fixing $\alpha/\beta$ will not change the performance very much. 

\subsection{Robustness Against Structural Perturbations (RQ4)}\label{sec:exp-robustness}
\begin{figure}[t]
    \centering
    \includegraphics[width = 0.8\linewidth]{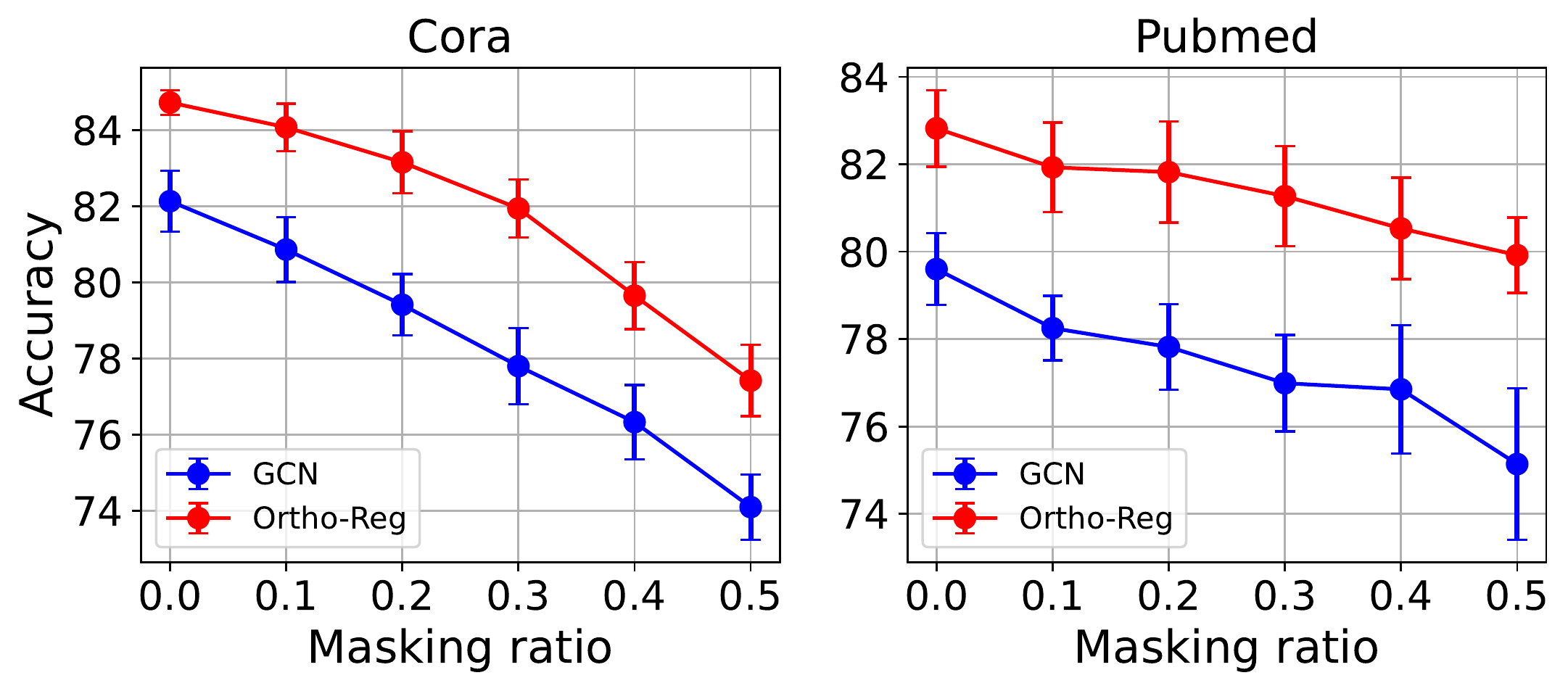}
    \caption{Effects of increasing of edge masking ratios.}
    \label{fig:exp-robustness}
\end{figure}

Finally, we study the robustness of \modelname against attacks on the graph structures compared with GNN models. As \modelname uses node features rather than a combination of node features and edges for prediction, we expect it to demonstrate better robustness under mild structural perturbations. To reach this target, we randomly mask a fraction of the edges of the graph and evaluate the performance of \modelname and GCN under different edge-masking ratios. In Fig.~\ref{fig:exp-robustness}, we plot how the model's performance changes (with standard deviation) as the masking ratio increases with 20 random trials.

As demonstrated in Fig.~\ref{fig:exp-robustness}, our method demonstrates better robustness against moderate-level edge perturbations. This is because we do not explicitly use the graph structure for generating predictions, making \modelname less sensitive to perturbations on the graph structure.

\section{Conclusions}\label{sec:conclusion}
In this paper, we have proposed \modelname, a novel Graph-Regularized MLP method for node representation learning. We show that simple graph regularization methods can cause dimensionally collapsed node embeddings both theoretically and empirically. We show that the proposed \modelname, which enforces the orthogonality of the correlation matrix of node embeddings, can naturally avoid the feature collapse phenomenon. We have conducted extensive experiments, including traditional transductive semi-supervised node classification tasks and inductive node classification for cold-start nodes, demonstrating the superiority of \modelname.

% \section*{Acknowledgements}

% \textbf{Do not} include acknowledgements in the initial version of
% the paper submitted for blind review.

% If a paper is accepted, the final camera-ready version can (and
% probably should) include acknowledgements. In this case, please
% place such acknowledgements in an unnumbered section at the
% end of the paper. Typically, this will include thanks to reviewers
% who gave useful comments, to colleagues who contributed to the ideas,
% and to funding agencies and corporate sponsors that provided financial


{
\bibliographystyle{icml2023}
\bibliography{ref}
}
%%%%%%%%%%%%%%%%%%%%%%%%%%%%%%%%%%%%%%%%%%%%%%%%%%%%%%%%%%%%%%%%%%%%%%%%%%%%%%%
%%%%%%%%%%%%%%%%%%%%%%%%%%%%%%%%%%%%%%%%%%%%%%%%%%%%%%%%%%%%%%%%%%%%%%%%%%%%%%%
% APPENDIX
%%%%%%%%%%%%%%%%%%%%%%%%%%%%%%%%%%%%%%%%%%%%%%%%%%%%%%%%%%%%%%%%%%%%%%%%%%%%%%%
%%%%%%%%%%%%%%%%%%%%%%%%%%%%%%%%%%%%%%%%%%%%%%%%%%%%%%%%%%%%%%%%%%%%%%%%%%%%%%%
\newpage
\appendix
\onecolumn

\section{Proofs}
\subsection{Proofs for Lemma~\ref{lemma:weight_matrix-shrink}}\label{proof:weight_matrix-shrink}
\begin{proof}
    First, let's take the gradient of the regularization loss $\gL_{reg}$ with respect to the weight matrix $\mW$:
    \begin{equation}
    \begin{split}
        \frac{{\partial} \gL_{reg}}{{\partial } \mW} = & \frac{\partial \; \text{tr} (\mH^{\top}\mL \mH)}{\partial \mW} \\
        = & \frac{\partial \; \text{tr} ((\mX\mW)^{\top}\mL (\mX\mW))}{\partial \mW} \\
        = & 2\mX^{\top}\mL\mX\mW \\
        = & 2\mP \mW
    \end{split}
    \end{equation}
    Treat the weight matrix as a function of the training step $t$, i.e., $\mW = \mW(t)$, then we can derive the gradient of $\mW(t)$ with respect to $t$ by $\frac{{\rm d}\mW(t)}{{\rm d} t} = 2\mP\mW$. As both $\mX$ and $\mL$ are fixed, we can solve the equation analytically,
    \begin{equation}
        \mW(t) = \exp (\mP t)\cdot \mW(0).
    \end{equation}
    As we have the non-ascending eigenvalues of $\mP$ as $\lambda_1^{\mP} \ge \lambda_2^{\mP} \ge \cdots \ge \lambda_D^{\mP}$, we can
    define an auxiliary function $f(t;\lambda_i^{\mP}, \lambda_j^{\mP}) = \exp(\lambda_i^{\mP}t)/\exp(\lambda_j^{\mP}t) = e^{(\lambda_i^{\mP} - \lambda_j^{\mP})t}$. It is obvious that $f(t;\lambda_i^{\mP}, \lambda_j^{\mP})$ is monotonically decreasing for all $j \ge i$.  As $\mW(t)$ is a transformation of its initial state $\mW(0)$ up to $\exp(\mP t)$, we can easily conclude that
    \begin{equation}
        \frac{\sigma^{\mW}_i(t)}{\sigma^{\mW}_j(t)} \le \frac{\sigma^{\mW}_i(t')}{\sigma^{\mW}_j(t')}, \;   \; \forall \; t < t' \; \mbox{and} \; i \le j. 
    \end{equation}

    If we further have the condition that $\lambda^{\mP}_1 \ge \cdots \ge \lambda^{\mP}_d > \lambda^{\mP}_{d+1} \ge \cdots \ge \lambda^{\mP}_D $, we have 
    $\lim\limits_{t\rightarrow \infty} f(t;\lambda_i^{\mP}, \lambda_j^{\mP}) = 0, \forall i \le d, j \ge d+1$. Then we are able to complete the proof.
\end{proof}
\subsection{Proofs for Theorem~\ref{theorem:dimensional-collapse}}\label{proof:dimensional-collapse}
\begin{proof}
    The embedding space is identified by the eigenspectrum of the correlation (covariance) matrix $\mC$ of node embeddings $\mH$. As $\mH = \mX \mW$, its correlation matrix can be (simply) identified as:
    \begin{equation}
    \begin{split}
        \mC = & \sum\limits_{i=1}^N (\vh_i - \overline{\vh})^{\top} (\vh_i - \overline{\vh}) / N \\
            = & \sum\limits_{i=1}^N \mW^{\top}(\vx_i - \overline{\vx})^{\top}(\vx_i - \overline{\vx})\mW / N. \\
    \end{split}
    \end{equation}
    According to Lemma ~\ref{lemma:weight_matrix-shrink}, $\mW$ has shrinking singular values, so $\mC$ has vanishing eigenvalues, indicating collapsed dimensions.

    Specially, when the input features have an identity matrix, we have:
    \begin{equation}
    \begin{split}
        \mC = & \sum\limits_{i=1}^N \mW^{\top}(\vx_i - \overline{\vx})^{\top} (\vx_i - \overline{\vx})\mW / N. \\
        = & \mW^{\top} \sum\limits_{i=1}^N \frac{(\vx_i - \overline{\vx})^{\top}(\vx_i - \overline{\vx})}{N} \mW \\
        = & \mW^{\top}\mW.
    \end{split}
    \end{equation}
\end{proof}
Thus, for $\mC$'s eigenvalues $\{\lambda_i^{\mC}\}_{i=1}^{D}$, we have $\lambda_i^{\mC} = (\sigma_i^{\mW})^2$. Then Theorem~\ref{theorem:dimensional-collapse} can be easily concluded with Lemma~\ref{lemma:weight_matrix-shrink}.
\subsection{Proofs for Theorem~\ref{theorem:ortho-reg}}\label{proof:ortho-reg}
\begin{proof}
    Note that we aim to optimize the following objective function (Eq.~\ref{eqn:ortho-reg}):
    \begin{equation*}
        \mathcal{L} = -\alpha \sum\limits_{k=1}^{D} C_{kk} + \beta \sum\limits_{k\neq k'}C_{kk'}^2. 
    \end{equation*}
As the first term applies only on the on-diagonal terms of the correlation matrix and the second term applies only on the off-diagonal terms, we are able to study the effects of two terms respectively:
\begin{equation}
    \begin{split}
        \mathcal{L}_{on-diag} & = -\alpha \sum\limits_{k=1}^D C_{kk} \\
         \mathcal{L}_{off-diag} & =  \beta \sum\limits_{k \neq k'}^D C_{kk'}^2 \\
    \end{split}
\end{equation}
For the on-diagonal terms $\gL_{on-diag}$, as we have set $T = 1$, we have $\mC = \sum\limits_{i=1}^{N} \vh_i \vs_i^{\top} / N$, and $C_{kk} = \sum\limits_{i=1}^{N} (h_i)_k \cdot (s_i)_k / N$ (the subscript $k$ denotes the $k$-th dimension). Then,
\begin{equation}
\begin{split}
    \frac{\partial C_{kk}}{\partial (h_i)_k} = \frac{1}{N} (s_i)_k, \; \mbox{and} \; \frac{\partial C_{kk}}{\partial (h_i)_k'} = 0, \; \; \forall k' \neq k.
\end{split}
\end{equation}
As a result,
\begin{equation}\label{eqn:on-diag-smooth}
    \frac{\partial \sum\limits_{k = 1}^D C_{kk}}{\partial \vh_i} = \frac{1}{N}\vs_i = \frac{1}{N} \frac{\sum_{j\in \gN (i)}\vh_j}{|\mathcal{N}(i)|}.
\end{equation}
Eq.~\ref{eqn:on-diag-smooth} indicates that the on-diagonal terms force each node embedding to be smoothed within its first-order neighborhoods.

Then we turn to the off-diagonal terms $\gL_{off-diag}$. Similarity, we have $C_{kk'} = \sum\limits_{i=1}^N (h_i)_k \cdot (s_i)_{k'} / N$. As for both $\mH$ and $\mS$, the diagonal term of their correlation matrixes for each dimension should be equal to $1$, formally,
\begin{equation}
    \frac{\sum\limits_{i=1}^N (h_i)_k^2}{N} = \frac{\sum\limits_{i=1}^N (s_i)_k^2}{N} = 1.
\end{equation}
Then according to Cauchy–Schwarz inequality, we have:
\begin{equation}
\begin{split}
    \left[\sum\limits_{i=1}^N (h_i)_k \cdot (s_i)_k\right]^2 & \le  \left[\sum\limits_{i=1}^N (h_i)_k^2\right]\left[\sum\limits_{i=1}^N (s_i)_k^2\right] = N^2 \\
    \frac{\sum\limits_{i=1}^N (h_i)_k \cdot (s_i)_k}{N} & \le 1 \\
\end{split},
\end{equation}
and the equality holds if and only if $(h_i)_k = (s_i)_k, \forall i$. As a result, the global optimizer of $\gL_{on-diag} $ will induce $\vh_i = \vs_i, \forall i$. Then, when $C_{kk'} = 0, k\neq k'$, we have:
\begin{equation}
    \sum\limits_{i=1}^N (h_i)_k \cdot (s_i)_{k'} / N = \sum\limits_{i=1}^N  (h_i)_k \cdot (h_i)_{k'} / N = C^{auto}_{kk'} = 0,
\end{equation}
where $\mC^{auto}$ is the auto-correlation matrix of $\mH$. So we've completed the proof.
\end{proof}

\section{Experiment Details in Section~\ref{sec:experiments}}\label{appendix-exp-detail}
\subsection{Datasets Details}
We present the statistics of datasets used in transductive node classification tasks in Table~\ref{tbl-statistics-trans}, including the number of nodes, number of edges, number of classes, number of input node features as well as the number of training/validation/testing nodes.
\begin{table}[h]
	\centering
	\caption{Statistics of benchmarking datasets in transductive settings}
	\label{tbl-statistics-trans}
	\small
	\begin{threeparttable}
	\setlength{\tabcolsep}{4mm}{
		\begin{tabular}{llllll}
			\toprule[0.8pt]
			Dataset  & \#Nodes & \#Edges & \#Classes & \#Features & \#Train\slash Val\slash Test \\
			\midrule[0.6pt]
            \cora     & 2,708   &  10,556 &  7 & 1,433 & 140/500/1000\\ 
            \citeseer & 3,327   &  9,228  & 6  & 3,703 & 120/500/1000 \\
            \pubmed   & 19,717  &  88,651 & 3  & 500 & 60/500/1000 \\
            \cs   & 18,333  &  327,576  & 15  & 6,805  & 300/500/1000 \\
            \physics  & 34,493  &  991,848  & 5  & 8,451  & 100/500/1000 \\
		  \computer & 13,752 & 574,418 & 10 & 767  & 200/500/1000\\
		  \photo   & 7,650 & 287,326 & 8  & 745  & 160/500/1000 \\
			\bottomrule[0.8pt]
		\end{tabular}}
	\end{threeparttable}
\end{table}

For the inductive node classification tasks in cold-start settings, we also present the statistics of \cora, \citeseer and \pubmed in Table~\ref{tbl-statistics-induc}, where we provide the number of isolated nodes, the number of tail nodes as well as the number of edges left after removing the isolated nodes from the graph.
\begin{table}[h]
	\centering
	\caption{Statistics of benchmarking datasets in inductive settings}
	\label{tbl-statistics-induc}
	\small
	\begin{threeparttable}
	\setlength{\tabcolsep}{4mm}{
		\begin{tabular}{lll|lll}
			\toprule[0.8pt]
			Dataset  & \#Nodes & \#Edges & \#Isolated & \#Tail &  \#Edges left \\
			\midrule[0.6pt]
            \cora    & 2,708 & 10,556 & 534   &  534 & 9,516 \\ 
            \citeseer & 3,327   &  9,228  & 676  & 676 & 7,968  \\
            \pubmed   & 19,717  &  88,651 & 4,547 & 4,547 & 79,557  \\
			\bottomrule[0.8pt]
		\end{tabular}}
	\end{threeparttable}
\end{table}

\subsection{Details of Experiments on Cold-Start Scenarios (Sec.~\ref{exp:inductive})}\label{appendix-exp-inductive}
In this section, we elaborate detailedly how the inductive isolated nodes are selected. Note that our processing directly follows the officially-implemented codes\footnote{\url{https://github.com/amazon-research/gnn-tail-generalization}} in ColdBrew~\citep{coldbrew}.
For each dataset, we first rank among the nodes according to their node degrees, through which we are able to get the degree of the bottom 3th percentile node, termed by $d_{3th}$. Then we screen out nodes whose degree is smaller than or equal to $d_{3th}$ as isolated nodes, which are subsequently removed from the original graph.

Note that in these datasets, most of the nodes only have a few connections (e.g. 1 or 2), the actual numbers of isolated nodes and tail nodes are usually much larger than the expected $3\%$. See Table~\ref{tbl-statistics-induc} for details.

For each dataset, we use the fixed 20 nodes per class (as in the public split) for training and all the remaining nodes for testing.

\subsection{Introduction of baselines} \label{baselines}
In Sec.~\ref{sec:exp-steups} we've briefly introduced the baselines for comparison, here we'd like to detailedly introduce the MLP-based baselines.

\paragraph{KD-MLPs:} We have covered two KD-MLP models: GLNN~\citep{glnn} and ColdBrew~\citep{coldbrew}. 
\begin{itemize}
    \item \textbf{GLNN}: As a typical GNN-to-MLP knowledge distillation method, given the predicted soft labels from a well-learned GNN model $\{\bm{z}_i\}$, GLNN learns an MLP model through jointly optimizing the supervised loss on labeled nodes and the cross-entropy loss between MLP's predictions and GNN's predictions over all nodes:
\begin{equation}
\begin{split}
    \mathcal{L}_{glnn} = & \mathcal{L}_{sup} + \lambda \mathcal{L}_{kd} \\
    \mathcal{L}_{sup} = \sum\limits_{i\in\mathcal V^L} \bm{\ell}_{xent} (\vy, \hat{\vy})  \;
 \mbox{and} & \; \mathcal{L}_{{kd}} = \sum\limits_{i\in\mathcal{V}} {\gD_{KL}} (\hat{\bm{y}}_i, \bm{z}_i),   \\
\end{split}
\end{equation}
where $\lambda$ is a trade-off hyperparameter.

\item \textbf{ColdBrew}: As another KD-MLP model, ColdBrew is specially designed to handle cold-start problems and has a totally different formulation compared with GLNN. First, it equips the teacher GNN model with structural embedding so that it can overcome the oversmoothing issue. Then, besides the knowledge distillation loss, it discovers the virtual neighborhood of each node using the embedding learned from the student MLP. With this operation, the model is able to estimate the possible neighbors of each node and thus can generalize better in inductive cold-start settings.
\end{itemize}

\paragraph{GR-MLPs:} We then introduce the covered GR-MLP models, including Lap-Reg~\citep{label-propagation,lap-reg}, P-Reg~\citep{p-reg} and GraphMLP~\citep{graphmlp} detailedly. Besides the basic supervised cross-entropy loss, GR-MLPs employ a variety of regularization losses to inject the graph structure knowledge into the learning of MLPs implicitly.

\begin{itemize}
    \item \textbf{Lap-Reg}: Based on the graph homophily assumption, Lap-Reg enforces Laplacian smoothing on the predicted node signals over the graph structure. Its regularization target could be formulated as:
    \begin{equation}
        \gL_{lap-reg} = {\rm tr}(\mY^{\top}\mL \mY),
    \end{equation}
    where $\mY$ the predicted node signals and $\mL$ is the Laplacian matrix of the graph.
    \item \textbf{P-Reg}: Similar to Lap-Reg, P-Reg is also built on top of the graph homophily assumption. However, instead of using edge-centric smoothing regularization, P-Reg employs a node-centric proximity preserving term that maximizes the similarity of each node and the average of its neighbors. The regularization objective could be formulated as:
    \begin{equation}
        \gL_{P-reg} = \frac{1}{N}\phi(\mH, \tilde{\mA}\mH),
    \end{equation}
    where $\tilde{\mA}\mH$ is the propagated node embedding matrix, and $\phi$ is a function that measures the difference between $\mH$ and $\tilde{\mA}\mH$, which could be implemented with a variety of measures like Square Error, Cross Entropy, Kullback-Leibler Divergence, etc.
    \item \textbf{Graph-MLP}: Inspired by the success of contrastive learning, Graph-MLP tries to get rid of GNN models by contrasting between connected nodes. Formally:
    \begin{equation}
        \gL_{graph-mlp} = \frac{1}{N}\sum\limits_{i=1}^N -\log\frac{\sum \limits_{j \in \mathcal{N} (i)} \exp({\rm sim}(\vh_i, \vh_j) /\tau )}{\sum \limits_{k \in \mathcal{V}} \exp({\rm sim}(\vh_i, \vh_k) /\tau )}.
    \end{equation}
    The final objective function is also a trade of between the supervised cross-entropy loss and the regularization loss.
\end{itemize}

\section{Additional Experiments}\label{appendix-exp-add}
\subsection{Experiments on OGB-Graphs}\label{appendix-exp-ogb}
To study the effectiveness of \modelname on large-scale graphs, we conduct experiments on two large-scale graphs: \arxiv and \products, and we present the results in this section.

The statistics of the two datasets in the transductive setting and the inductive cold-start setting are presented in Table~\ref{tbl-statistics-trans-ogb} and Table~\ref{tbl-statistics-induc-ogb}. Note that the official split of OGB datasets is different from other datasets in Sec.~\ref{sec:experiments} and does not follow the semi-supervised setting.

\begin{table}[h]
	\centering
	\caption{Statistics of OGB datasets in transductive settings}
	\label{tbl-statistics-trans-ogb}
	\small
	\begin{threeparttable}
        {
		\begin{tabular}{llllll}
			\toprule[0.8pt]
			Dataset  & \#Nodes & \#Edges & \#Classes & \#Features & \#Train\slash Val\slash Test \\
			\midrule[0.6pt]
            \arxiv     & 169,343   &  2,332,486 &  40 & 128 & 90,941 / 29,799 / 48,603\\ 
            \products & 2,449,029   &  123,718,024  & 47  & 100 & 196,615 / 39,323 / 2,213,091  \\
			\bottomrule[0.8pt]
		\end{tabular}}
	\end{threeparttable}
\end{table}

\begin{table}[h]
	\centering
	\caption{Statistics of OGB datasets in inductive settings}
	\label{tbl-statistics-induc-ogb}
	\small
	\begin{threeparttable}
        {
		\begin{tabular}{lll|lll}
			\toprule[0.8pt]
			Dataset  & \#Nodes & \#Edges  & \#Isolated & \# Nodes left &\# Edges left\\
			\midrule[0.6pt]
            \arxiv     & 169,343   &  2,332,486 & 16,934 & 152,409 & 2,298,618 \\ 
            \products & 2,449,029 & 123,718,024 & 244,902  & 2,204,127 & 123,661,058    \\
			\bottomrule[0.8pt]
		\end{tabular}}
	\end{threeparttable}
\end{table}

% We present the performance of \modelname in transductive node classification setting on OGB datasets in Table~\ref{tbl-statistics-trans-ogb} and Table~\ref{tbl-statistics-induc-ogb}.

For fair comparison, we use the same model size (i.e., the number of parameters) for each model. We train each model for 10 times on each dataset and report the average accuracy with standard deviation on transductive setting and inductive cold-start setting in Table~\ref{tbl-exp-trans-ogb} and Table~\ref{tbl-exp-induc-ogb} respectively.

\begin{figure}[h]
    \begin{minipage}[h]{0.48\linewidth}
    \vspace{0pt}
    \centering
    \captionof{table}{Test accuracy on OGB datasets in transductive settings.}
	\label{tbl-exp-trans-ogb}
	\small
     \setlength{\tabcolsep}{0.8mm}
    {   
	\begin{threeparttable}
        {
        \scalebox{0.80}
        {
		\begin{tabular}{clccccccc}
			\toprule[0.8pt]
		 & Methods & \arxiv & \products \\
			\midrule
	\multirow{2}{*}{GNNs} & GCN &  \textbf{71.74}$\pm$\textbf{0.29} & 75.26$\pm$0.21  \\
          & SAGE &  71.49$\pm$0.27 & \textbf{78.61}$\pm$\textbf{0.23}  \\
      	\midrule[0.5pt]
        KD-MLPs & GLNN &  69.37$\pm$0.25 & {75.19$\pm$0.34}  \\
          \midrule[0.5pt]
         \multirow{4}{*}{GR-MLPs} & MLP & 56.28$\pm$0.37 & 61.06$\pm$0.08  \\
  	 & Lap-Reg & 57.83$\pm$0.52 & 65.91$\pm$0.31 \\
  	 & P-Reg & 58.41$\pm$0.45 & 65.32$\pm$0.28  \\
  	 & GraphMLP & 61.11$\pm$0.36 & 68.54$\pm$0.33  \\
                \midrule[0.5pt]
  	Ours &  \modelname &  {70.35$\pm$0.22} & 74.35$\pm$0.19  \\
	   \bottomrule[0.8pt] 
		\end{tabular}
          }
          }
	\end{threeparttable}
	}
    \end{minipage}
    \begin{minipage}[h]{0.5\linewidth}
    \vspace{0pt}
    \centering
    \captionof{table}{Test accuracy on the isolated nodes of OGB datasets.}
	\label{tbl-exp-induc-ogb}
	\small
    \setlength{\tabcolsep}{0.75mm}
    {   
	\begin{threeparttable}
        {
        \scalebox{0.78}
        {
	\begin{tabular}{llcccc}
	\toprule[0.8pt]
		  \multicolumn{2}{c}{{Methods}} & \arxiv & \products  \\
	\midrule
		 \multirow{2}{*}{{GNNs}} & GCN  & 44.51$\pm$0.85 & 56.62$\pm$1.12  \\
          & GraphSAGE & 47.32$\pm$0.89 & 57.88$\pm$1.01   \\
          \specialrule{0em}{1pt}{1pt}
          \midrule
          \specialrule{0em}{1pt}{1pt}
         \multirow{2}{*}{{KD-MLPs}} & ColdBrew & 52.36$\pm$0.84 & 61.64$\pm$0.98    \\
          & GLNN & {53.18$\pm$1.05} & {63.09$\pm$0.87} \\
        \specialrule{0em}{1pt}{1pt}
          \midrule
          \specialrule{0em}{1pt}{1pt}
             \multirow{5}{*}{{GR-MLPs}} & MLP & 51.03$\pm$0.75 & 60.18$\pm$0.84  \\
             & Lap-Reg & 51.87$\pm$0.81 & 60.47$\pm$0.77 \\
             & P-Reg & 51.79$\pm$0.88 & 60.59$\pm$0.91 \\
             & GraphMLP & 52.21$\pm$0.91 & 61.12$\pm$0.98 \\
             & \modelname (Ours) & {\textbf{54.51}$\pm$\textbf{0.77}} & {\textbf{63.95}$\pm$\textbf{0.74}}  \\
        \bottomrule[0.8pt] 
	\end{tabular}}}
	\end{threeparttable}
	}
    \end{minipage}
\end{figure}
As demonstrated in Table~\ref{tbl-exp-trans-ogb}, though under-performing GNN models, \modelname gives a quite good performance (which is very close to that of GNNs) on these two datasets. On inductive cold-start prediction tasks, \modelname also outperforms both GNN models and other MLP models. 

\subsection{Experiments on Heterophily Graphs}\label{appendix-exp-hete}
Then we study the generalization ability of \modelname on heterophily (non-homophily) graphs. We first give the formal definition of graph homophily ratio as follows:
\begin{definition}\label{def:homo_ratio}
(Graph Homophily Ratio)
For a graph $\mathcal{G} = (\mathcal{V}, \mathcal{E})$ with adjacency matrix $\bm{A}$,  its  homophily ratio $\phi$ is defined as the probability that two connected nodes share the same label:
\begin{equation}\label{eqn:homo-ratio}
    \phi = \frac{\sum_{i,j \in \mathcal{V}} A_{ij} \cdot \mathbbm{1}[{\bm{y}_i = \bm{y}_{j}}]}{\sum_{i,j \in \mathcal{V}} A_{ij}} = \frac{\sum_{i,j \in \mathcal{V}} A_{ij} \cdot \mathbbm{1}[{\bm{y}_i = \bm{y}_{j}}]}{|\mathcal{E}|}
\end{equation}
\end{definition}

The evaluated datasets previously are all homophily graphs, where connected nodes tend to share the same labels. To evaluate \modelname more extensively we consider three more widely used heterophily graphs: \chameleon, \squirrel and \actor. We provide statistics of the three datasets in Table~\ref{tbl-statistics-hete}.
\begin{table}[h]
	\centering
	\caption{Statistics of heterophily graphs}
	\label{tbl-statistics-hete}
	\small
	\begin{threeparttable}
	{
		\begin{tabular}{llllll}
			\toprule[0.8pt]
			Dataset  & \#Nodes & \#Edges & \#Classes & \#Features & Heterophily ratio $\phi$ \\
			\midrule[0.6pt]
            \chameleon     & 2,277   &  36,101 & 5 & 2,325 & 0.25 \\ 
            \squirrel & 5,201   &  217,073  & 5  & 2,089 & 0.22  \\
            \actor & 7,600   &  33,544  & 5  & 931 & 0.24 \\
		\bottomrule[0.8pt]
		\end{tabular}}
	\end{threeparttable}
\end{table}

For heterophily graphs where the graph homophily assumption does not hold~\citep{homo, homo-citation}, it might be improper to enforce node embeddings/predictions to be smoothed over the graph structure. As a result, we modify the neighborhood abstraction function so that it can better adjust to heterophily graphs. Specifically, we adopt the following function to construct summary embedding:
\begin{equation}
    \mS = \tilde{A}^2 \mH / T.
\end{equation}
Compared with Eq.~\ref{eqn:neighbor-summary}, we use only the second neighbors for heterophily graphs. We present the results on the three heterophily graphs in Table~\ref{tbl-exp-hete}. For comparison with GNN models, we cover two additional GNN models that are specifically designed for handling heterophily graphs: Geom-GCN~\citep{geom-gcn} and GPRGNN~\citep{gprgnn}

\begin{table}[h]
	\centering
	\caption{Performance on heterophily graphs}
	\label{tbl-exp-hete}
	\small
	\begin{threeparttable}
        {
        \scalebox{1.0}
        {
		\begin{tabular}{clccc}
			\toprule[0.8pt]
		 & Methods & \chameleon & \squirrel & \actor \\
			\midrule
	\multirow{2}{*}{GNNs}
          & Geom-GCN &  67.32$\pm$1.76 & 46.01$\pm$1.27 & 30.59$\pm$0.76  \\
        & GPRGNN &  66.31$\pm$2.05 &   50.56$\pm$1.51 & 30.78$\pm$0.83  \\
      	\midrule[0.5pt]
        KD-MLPs & GLNN &  60.58$\pm$1.72 & 43.72$\pm$1.16 & 34.12$\pm$0.77  \\
          \midrule[0.5pt]
         \multirow{3}{*}{GR-MLPs} & MLP & 47.59$\pm$0.73 & 31.67$\pm$0.61 & 35.93$\pm$0.61  \\
  	 & Lap-Reg & 48.72$\pm$1.52 & 30.44$\pm$0.97 &  33.71$\pm$0.59 \\
  	 &  \modelname (ours)&  \underline{63.55$\pm$0.83} & \underline{48.72$\pm$0.93}  & \underline{\textbf{36.64}$\pm$\textbf{0.67}} \\
	   \bottomrule[0.8pt] 
		\end{tabular}
          }
          }
	\end{threeparttable}
\end{table}

As demonstrated in the Table, though unable to match the performance of advanced GNN models on heterophily graphs, our method greatly narrows the gap between MLPs and GNNs. Specifically, on heterophily graphs where GNNs even perform worse than the vanilla MLP, our model is able to achieve even better performance, thanks to the MLP-based encoder and the regularization loss.

% \subsection{Effect of the Strengths of Regularization}
% We also investigate the impacts of the strengths of the two regularization losses $\alpha$, $\beta$, on \modelname's performance. To reach this target, we try different $\alpha$, $\beta$ combinations and plot the model's performance on \cora, \citeseer and \pubmed in heatmaps. The reported results are averaged over 20 trials.

\subsection{Scalability Test}\label{appendix-exp-scale}
\begin{figure}[h]
    \centering
    \includegraphics[width = 0.5\linewidth]{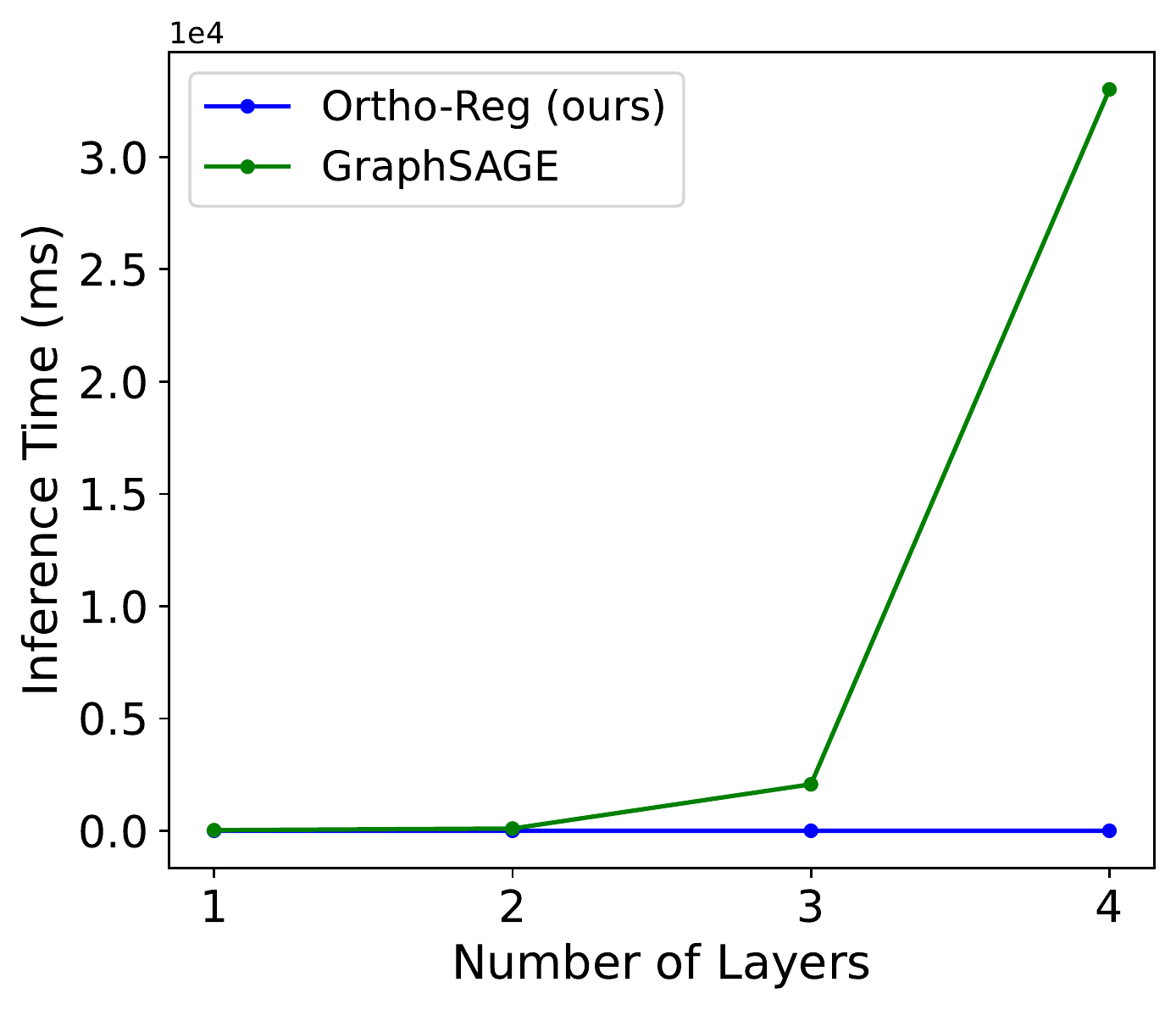}
    \caption{The inference time comparison of GraphSAGE and \modelname on 
    \products. \modelname is able to perform inference much faster than GraphSAGE.}
    \label{fig:exp-time}
\end{figure}

Finally, we show that with an MLP model as the encoder, \modelname is able to perform inference fast without the reliance on graph structure. We plot the inference time of \modelname and GraphSAGE on \products with different model depths in Fig.~\ref{fig:exp-time}. The result demonstrates the superiority of the inference benefit of \modelname over GNN models.

\subsection{Sensitivity analysis of trade-off hyperparameters}\label{appendix-exp-sense}
\begin{figure}[h]
    \centering
    \includegraphics[width = 1.0\linewidth]{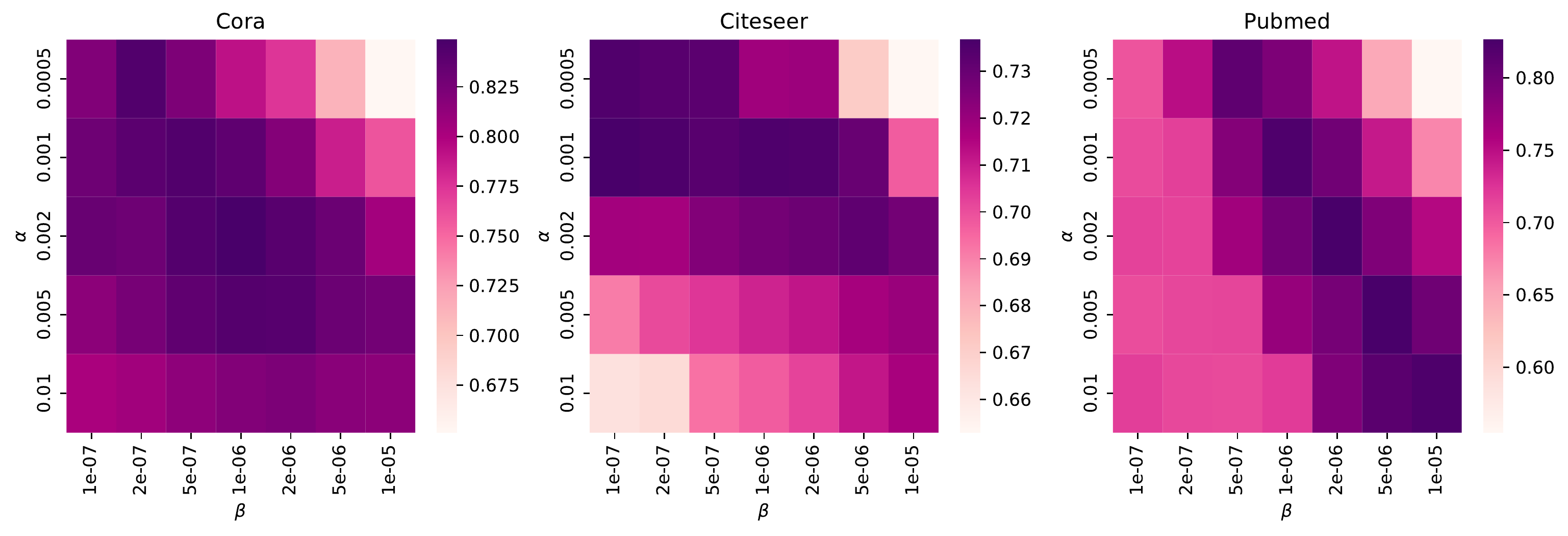}
    \caption{Performance heat map when using different $\alpha$, $\beta$ combinations in Eq.~\ref{eqn:ortho-reg}, on \cora, \citeseer and \pubmed.}
    \label{fig:sense-citeseer}
\end{figure}

% You can have as much text here as you want. The main body must be at most $8$ pages long.
% For the final version, one more page can be added.
% If you want, you can use an appendix like this one, even using the one-column format.
%%%%%%%%%%%%%%%%%%%%%%%%%%%%%%%%%%%%%%%%%%%%%%%%%%%%%%%%%%%%%%%%%%%%%%%%%%%%%%%
%%%%%%%%%%%%%%%%%%%%%%%%%%%%%%%%%%%%%%%%%%%%%%%%%%%%%%%%%%%%%%%%%%%%%%%%%%%%%%%

\end{document}